\documentclass[11pt]{article}

\usepackage[final]{acl}

\makeatletter
\@ifpackageloaded{lineno}{%
\@ifpackagelater{lineno}{2026/02/28}{}{%
  \RemoveFromHook{build/column/before}[lineno]%
  \AddToHook{build/column/after}{%
    \setbox\@outputbox\vbox{%
      \boxmaxdepth\@maxdepth
      \protected@write\@auxout{}{%
        \string\def\string\@LN@column{\if@firstcolumn1\else2\fi}%
      }%
      \unvbox\@outputbox
    }%
  }%
}%
}{}
\makeatother

\newlength{\abstractheadgap}   
\newlength{\abstractfootgap}   
\makeatletter
\renewenvironment{abstract}%
  {\setlength{\topsep}{\z@}%
   \setlength{\partopsep}{\z@}%
   \begin{center}\large\textbf{\abstractname}\end{center}%
   \vspace{\abstractheadgap}%
    \begin{list}{}%
      {\setlength{\rightmargin}{0.6cm}%
        \setlength{\leftmargin}{0.6cm}%
        \setlength{\topsep}{\z@}%
        \setlength{\partopsep}{\z@}%
        \setlength{\parsep}{\z@}}%
      \item[]\ignorespaces%
      \@setsize\normalsize{12pt}\xpt\@xpt
  }%
  {\unskip\end{list}\vspace{\abstractfootgap}}
\makeatother

\usepackage{times}
\usepackage{latexsym}
\usepackage{amssymb}
\usepackage{booktabs}
\usepackage{multirow}

\usepackage[T1]{fontenc}

\usepackage[utf8]{inputenc}

\usepackage{microtype}

\usepackage{inconsolata}

\usepackage{graphicx}

\usepackage{relsize}

\definecolor{cigray}{gray}{0.45}
\newcommand{\pmci}[1]{\kern0.1em\textcolor{cigray}{\raisebox{0.5ex}{\relscale{0.85}\textpm#1}}}

\usepackage{tikz}
\usetikzlibrary{positioning, arrows.meta, fit, calc, shapes.geometric, backgrounds}
\usepackage{xcolor}
\definecolor{osLatent}{RGB}{31, 78, 121}
\definecolor{osRender}{RGB}{0, 112, 92}
\definecolor{osArtifact}{RGB}{176, 78, 0}
\definecolor{osRefine}{RGB}{120, 47, 138}
\definecolor{osBgLatent}{RGB}{232, 240, 250}
\definecolor{osBgRender}{RGB}{226, 244, 238}
\definecolor{osBgArtifact}{RGB}{252, 235, 218}
\definecolor{osBgRefine}{RGB}{243, 232, 248}
\definecolor{osBgL1}{RGB}{245, 245, 245}
\definecolor{osBgL2}{RGB}{229, 235, 245}
\definecolor{osBgL3}{RGB}{206, 220, 240}

\title{OSynth: A Longitudinal, Context-Sensitive Evaluation of \\[1.25pt] Meeting Understanding}

\author{
  \bfseries Ian Porada, Suraj Tripathi, Chaitanya Animesh, Leda Sari,
  \\
  \bfseries Chaitanya Narisetty, Austin R. Matthews, Bohan Yu, Ramesh L. Thulasiram,
  \\
  \bfseries Gregory Sell, Kean Chin, Dmitriy Genzel
  \\
  \mdseries Otter.ai
  \\
  \mdseries Mountain View, California, USA
  \\
  \mdseries{
    \texttt{\{ian, stripathi, canimesh, kkchin, dmitriy\}@otter.ai}
  }
}

\begin{document}
\maketitle
\begin{abstract}
Meeting notetakers are increasingly deployed in settings where longitudinal context is available including calendar metadata, prior meetings, emails, slides, and speaker enrollments; however, meeting corpora remain limited in terms of size, domain coverage, and available context.
We present \textsc{OSynth}, a large-scale synthetic dataset of simulated organizations comprising 570 hours of meeting audio together with associated artifacts and a latent ground-truth state.
In a pairwise comparison, human raters distinguish \textsc{OSynth} meetings from real meeting transcripts with only 56\% accuracy based on excerpts of 150--300 words.
Across state-of-the-art LLMs, additional context yields significant gains on several entity-sensitive dimensions, demonstrating limitations of contemporary approaches and the potential usefulness of \textsc{OSynth} in developing better-contextualized meeting-understanding systems.
\end{abstract}

\section{Introduction}

Meeting understanding has expanded from summarization of an isolated transcript to a longitudinal problem in which one can draw on surrounding organizational artifacts such as prior meetings, shared documents, slide decks, internal emails, and speaker enrollments \citep{asthana-etal-2023-summaries, laskar-etal-2023-building, arakawa-etal-2025-meeting-minutes}. Existing meeting benchmarks were built largely before this shift: they consist of standalone sessions with limited surrounding contextual artifacts \citep{carletta-etal-2005-ami, janin-etal-2003-icsi, zhong-etal-2021-qmsum, hu-etal-2023-meetingbank, nedoluzhko-etal-2022-elitr}, and the community consequently has no way to measure, at scale, whether longitudinal grounding actually improves summarization or downstream outputs.

We introduce \textsc{OSynth},\footnote{\url{https://github.com/otter-ai/osynth}} a fully synthetic, longitudinally generated meeting corpus that closes this gap, and use it to show that as more context is made available, summary quality from state-of-the-art MLLMs improves significantly across entity-sensitive dimensions such as summary faithfulness, action-item assignee accuracy, and cross-meeting speaker-identity consistency.

\begin{figure}[t]
  \centering
  \includegraphics[width=1.0\columnwidth]{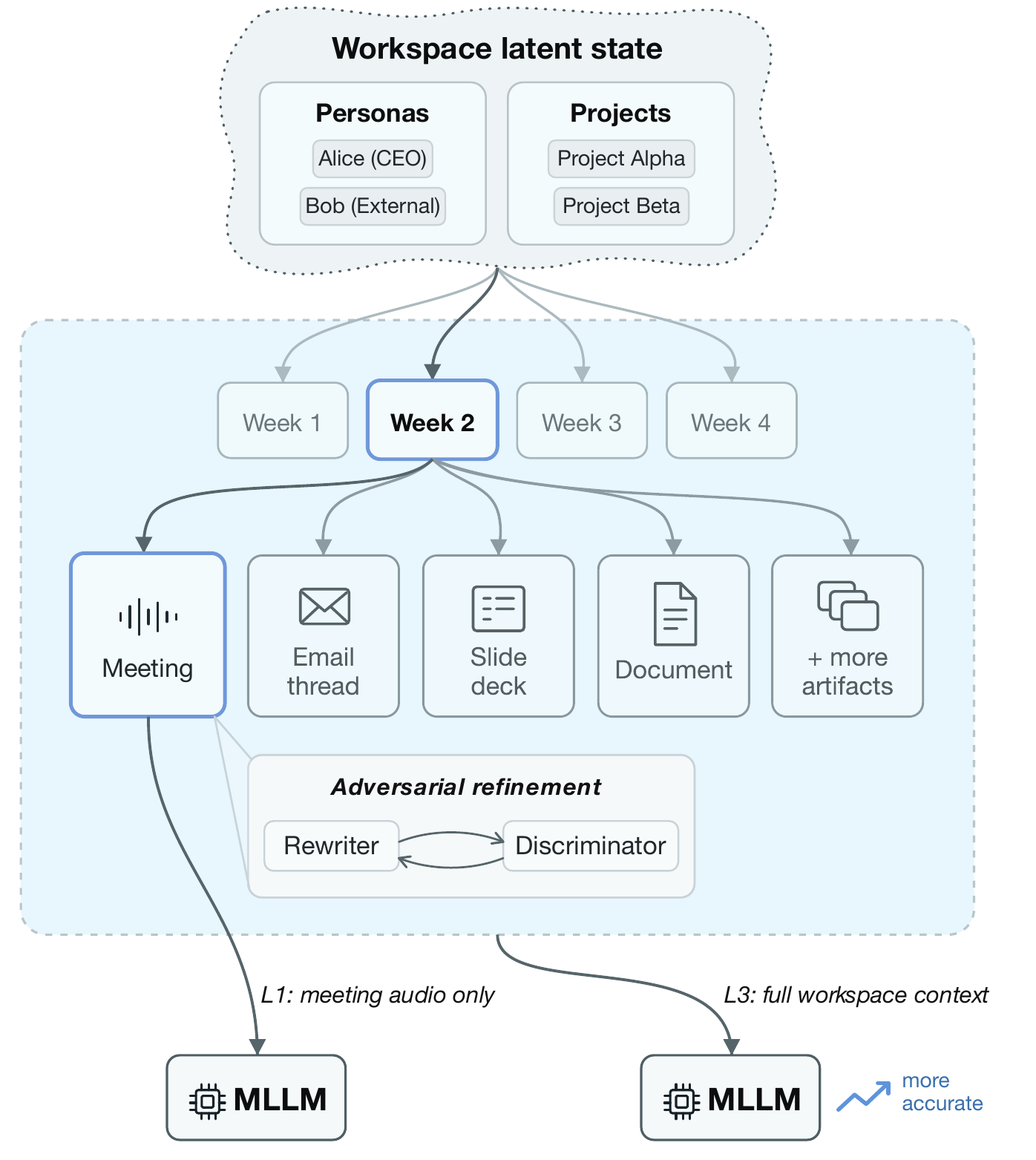}
  \caption{Each \textsc{OSynth} organization, termed a \textit{workspace}, is generated from a single shared latent state from which all artifacts (audio, transcripts, etc.) are derived. Meeting transcripts are adversarially refined to improve realism, and in experiments multimodal large language models (MLLMs) improve significantly in certain dimensions when given access to broader workspace context (L3) as compared to a single meeting's audio (L1).}
  \label{fig:teaser}
  \vspace{-8pt}
\end{figure}

Following existing work that uses language models to synthesize conversational data \citep{chen-etal-2023-places, kim-etal-2023-soda, ding-etal-2023-ultrachat, xu-etal-2023-baize}, \textsc{OSynth} is generated hierarchically by Claude Sonnet 4.6 \citep{anthropic-2026-claude-sonnet}. To ensure that the generated transcripts resemble real meetings, we develop an adversarial refinement process grounded in real AMI transcripts \citep{carletta-etal-2005-ami}, in which Claude Opus 4.7 \citep{anthropic-2026-claude-opus} refines drafts against Gemini 3.1 Pro \citep{google-2026-gemini-3-pro} as a realism discriminator (\S\ref{sec:dataset:refinement}). Each meeting's context maps to a \emph{workspace} which is a simulated organization with a latent state (personas, ongoing projects, prior decisions, outstanding tasks) from which every observable artifact is derived (Figure~\ref{fig:teaser}). Observable artifacts include: transcripts (turn-by-turn, with disfluencies), synthesized audio rendered with Qwen3-TTS \citep{qwen3-tts-2025} and simulated acoustics, executive summaries, action items, outlines, internal emails, and slide decks. 

Because the latent state is known by construction, evaluation metrics can be computed against an exact reference. \textsc{OSynth} also contains only synthetic data and is shareable in full. This is in contrast to published meeting corpora which must redact sensitive content---for instance, ICSI \citep{janin-etal-2003-icsi} anonymizes speaker identities and ELITR \citep{nedoluzhko-etal-2022-elitr} anonymizes named entities---or are relatively limited in size and scope. The final dataset spans 570 hours of audio across 160 workspaces, and the nature of the data creation pipeline allows for scaling to even larger sizes. The released corpus is English-only, although the pipeline is language agnostic, and we performed a preliminary data generation in five additional languages (Appendix~\ref{sec:appendix:multilingual}).

We use \textsc{OSynth} to evaluate two state-of-the-art systems (an OpenAI stack composed of GPT-4o and GPT-5.5 \citep{openai-2026-gpt-4o-transcribe-diarize, openai-2026-gpt-5-5}, and a Gemini stack that pairs Gemini 3.1 Pro \citep{google-2026-gemini-3-pro} for enrollment-based speaker identification with Gemini 3 Flash \citep{google-2025-gemini-3-flash} for transcription and summarization) under three progressively richer levels of context: (L1) single meeting audio only; (L2) L1 plus the model's own outputs on prior meetings; and (L3) full workspace artifacts (slides, emails, and speaker enrollments). Systems are scored both on isomorphic metrics derived directly from the latent state (atomic-fact coverage, named-entity spelling accuracy, and true-entity mention rate) and on standard summarization metrics such as LLM-as-a-judge \citep{zheng-etal-2023-judging}, which has been shown to underperform reference-grounded metrics in measuring factual accuracy \citep{tang-etal-2024-tofueval}.

We also validate the benchmark itself with a human-discrimination study (\S\ref{sec:dataset:realism}) in which annotators are shown a length-matched pair of one real and one synthetic excerpt and asked to identify the synthetic side; they do so significantly less often for \textsc{OSynth} (56.3\%) than for a control without our adversarial refinement (79.6\%; p-value 0.001).

In summary, our contributions are: (i) the \textsc{OSynth} pipeline and associated dataset, a novel pipeline for generating synthetic meetings with aligned workspace artifacts and an exact latent-state reference; (ii) a validation of the dataset's transcript-level realism through a pairwise human-discrimination study; and (iii) a context-sensitive evaluation of state-of-the-art, multimodal large language model (MLLM) families, showing that longitudinal context yields gains on entity-sensitive summarization and notetaking dimensions that existing benchmarks do not expose.

\section{Related work}

\begin{table*}[t]
  \centering
  \footnotesize
  \setlength{\tabcolsep}{5pt}
  \begin{tabular}{lrlp{2.8cm}p{5.5cm}}
    \toprule
    \textbf{Corpus} & \textbf{Hours} & \textbf{Domain} & \textbf{Reference} & \textbf{Contextual artifacts} \\
    \midrule
    ICSI           & 72          & Academic         & gold + recorded audio   & prior meetings (recurring groups) \\
    AMI            & 100         & Product design   & gold + recorded audio   & slides (subset), whiteboard video, prior meetings (scenario subset) \\
    AliMeeting     & 119         & Multi-domain (zh) & gold + recorded audio   & --- \\
    ELITR-Minuting & 80          & Technical        & gold, no audio          & minutes \\
    MeetingBank    & 3{,}579     & City council     & ASR + recorded audio    & agenda, prior meetings \\
    NOTSOFAR-1     & 32          & Office meetings  & gold + recorded audio   & --- \\
    \midrule
    \textsc{OSynth} (ours) & 570 & Multi-domain & gold + synthetic audio & calendar, prior meetings, emails, docs, slides, speaker-identity voice enrollments \\
    \bottomrule
  \end{tabular}
  \caption{\textsc{OSynth} compared to existing meeting corpora. \emph{Reference} encodes the transcription process (\emph{gold} vs.\ \emph{ASR}) and audio source (\emph{recorded} vs.\ \emph{synthetic}); \textsc{OSynth} is the only synthetic corpus in this table and is validated against real meeting transcripts by the human-discrimination study in \S\ref{sec:dataset:realism}.}
  \label{tab:benchmark-comparison}
\end{table*}

\paragraph{Meeting and dialogue corpora.}
Spoken meeting corpora such as AMI \citep{carletta-etal-2005-ami}, ICSI \citep{janin-etal-2003-icsi}, MeetingBank \citep{hu-etal-2023-meetingbank}, ELITR \citep{nedoluzhko-etal-2022-elitr}, AliMeeting \citep{yu-etal-2022-m2met}, and NOTSOFAR-1 \citep{vinnikov-etal-2024-notsofar} provide important testbeds, but they are either narrow in domain, lack audio, lack longitudinal workspace artifacts, or focus on ASR/diarization rather than summarization and downstream meeting understanding (Table~\ref{tab:benchmark-comparison}). The CHiME shared-task series \citep{barker-etal-2018-chime5, cornell-etal-2023-chime7-dasr} has driven progress on far-field ASR, diarization, and SID for meeting-style audio, but is upstream of the tasks we study such as summarization. The AutoMin shared-task series \citep{ghosal-etal-2021-automin, ghosal-etal-2023-overview, shinde-etal-2025-automin} instead targets the generation side directly, eliciting minutes from ELITR project meetings and European Parliament sessions and, in its third edition, question answering over meeting transcripts; participating methods moved from segment-then-summarize sequence-to-sequence pipelines to prompted LLMs. In both tasks the system input is the transcript of a single session, so the evaluation cannot isolate the effect of surrounding workspace context---the comparison \textsc{OSynth} is built to support. Dialogue-summarization resources such as QMSum \citep{zhong-etal-2021-qmsum}, SAMSum \citep{gliwa-etal-2019-samsum}, and MediaSum \citep{zhu-etal-2021-mediasum} broaden task format and domain, but do not expose the deployed workspace setting. \textsc{OSynth}, by contrast, combines longitudinal workspace context with an exact latent reference at a larger scale.

\paragraph{Evaluating context-augmented dialogue summarization.}
Prior work shows both the need for and the difficulty of evaluating context-aware summarization: dialogue summaries often hallucinate \citep{tang-etal-2024-tofueval}, reference-free LLM judges can underweight factual precision \citep{liu-etal-2023-revisiting}, and long-context benchmarks \citep{bai-etal-2024-longbench, an-etal-2024-leval} do not isolate longitudinal workspace grounding. Retrieval-augmented meeting systems \citep{asthana-etal-2023-summaries, laskar-etal-2023-building, arakawa-etal-2025-meeting-minutes} therefore require benchmarks that can score whether added context improves factual outputs. This motivates our use of \emph{isomorphic} latent-state metrics alongside conventional LLM-as-judge \citep{zheng-etal-2023-judging, liu-etal-2023-g-eval} scores.

\paragraph{Synthetic conversational data.}
LLM-prompted dialogue synthesis \citep{chen-etal-2023-places}, role-play and game-based corpora \citep{rameshkumar-bailey-2020-critical-role, collaborative-game-2025}, and agent simulations with persistent memory \citep{park-etal-2023-generative-agents, li-etal-2023-camel, hong-etal-2024-metagpt} show that synthetic multi-party interaction can scale. \textsc{OSynth} adapts this idea to meeting evaluation by generating not only conversations but also the surrounding workspace artifacts and latent references needed for scoring. Related to our transcript generation are computational models of multi-party turn-taking dynamics \citep{obryan-etal-2024-mlspeak}.

\section{The \textsc{OSynth} dataset}
\label{sec:dataset}

\begin{figure*}[t]
  \centering
  \includegraphics[width=1.0\textwidth]{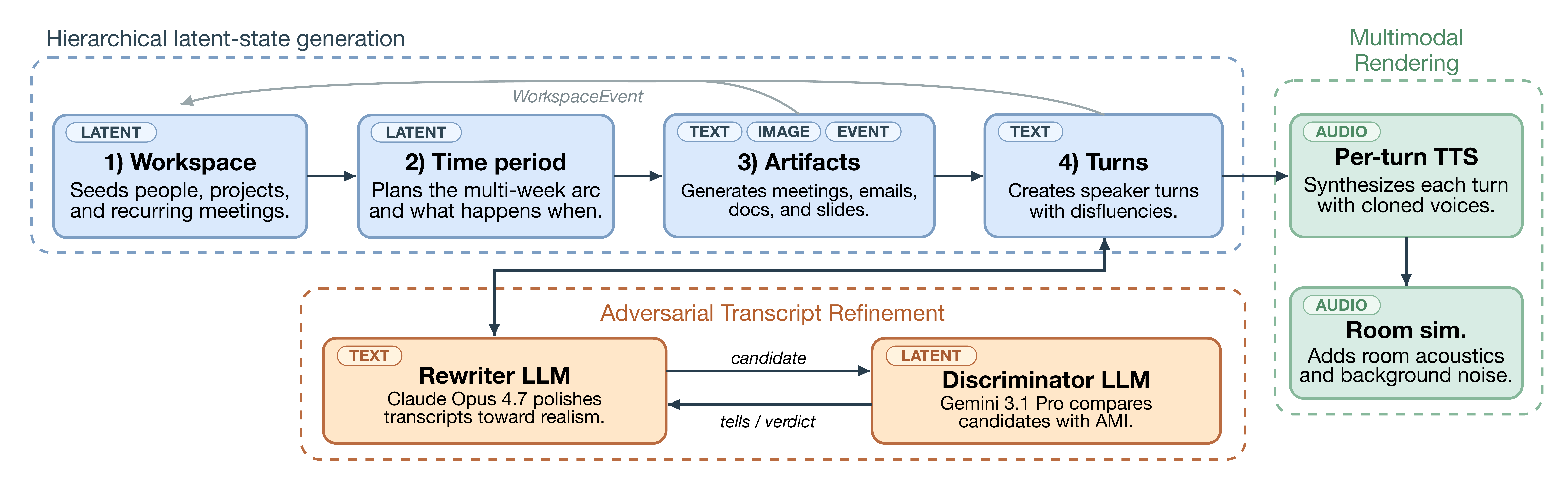}
  \caption{Overview of the \textsc{OSynth} synthetic data pipeline. A hierarchical LLM pipeline samples a workspace-level latent state, expands it into meetings and surrounding artifacts, renders the artifacts into audio and text, and maintains a rolling \textsc{WorkspaceState} so that every observable can be scored against an exact reference. Artifacts emit \textsc{WorkspaceEvents} that are considered when generating subsequent artifacts. Audio is synthesized turn-by-turn with voice cloning and the final audio channels are simualted (virtual, hybrid, or in-person acoustics) using \texttt{pyroomacoustics} \citep{scheibler-etal-2018-pyroomacoustics}.}
  \label{fig:pipeline}
  \vspace{-4pt}
\end{figure*}

\textsc{OSynth} is a corpus of fully synthetic workspaces. Each workspace is a simulated organization observed over a multi-week window, and is produced by a hierarchical LLM pipeline (\S\ref{sec:dataset:pipeline}, Figure~\ref{fig:pipeline}) that first samples a latent state and then derives every observable artifact from that state. The latent state comprises an org chart with personas and speaking styles, ongoing projects, a narrative arc of decisions and outstanding action items, and a calendar of recurring and ad-hoc meetings. The derived artifacts (\S\ref{sec:dataset:rendering}) include clean turn-by-turn transcripts, ASR-ready multi-speaker audio rendered via TTS with voice cloning, per-meeting summaries, action items, outlines, inter-meeting emails and shared documents, slide decks, and per-speaker voice enrollments. Because every artifact is generated from the same explicit latent state, each observable can be scored against an exact reference. We conduct a two-alternative forced choice (2AFC) human-discrimination study which confirms that \textsc{OSynth} transcripts are difficult to distinguish from real meeting transcripts (\S\ref{sec:dataset:realism}).

\subsection{Hierarchical generation pipeline}
\label{sec:dataset:pipeline}

The pipeline expands a workspace top-down across four levels of abstraction, with each level consuming the latent state produced by the previous one. We use \texttt{claude-sonnet-4-6} as the instruction-following LLM at every stage and pass a structured state representation through prompts to keep entities, decisions, and outstanding actions consistent across meetings. Appendix~\ref{sec:appendix:generation-details} gives additional implementation details.

\paragraph{1) Workspace.} The first level samples a self-contained organization: a company description (industry, size, stage), an org chart with named personas, per-persona attributes (role, seniority, communication style, demographic metadata used later for voice assignment), an inventory of active projects, and a baseline schedule of recurring meetings (standups, reviews, 1:1s). Names and organizations are sampled from a generated pool rather than reused from real entities to avoid real instances of PII and ensure data diversity.

\paragraph{2) Time period and narrative arc.} The second level fixes a simulated calendar window and writes weekly plans: which projects make progress, which decisions are made, which action items are opened and closed, and which external events intrude.

\paragraph{3) Artifacts.} The third level decomposes weekly plans into a concrete calendar of meetings together with the inter-meeting artifacts the same latent state drives: emails, shared documents, and slide decks tied to projects and meeting agendas. Each meeting spec is assigned an invite, agenda, attendees, and the slice of latent state it advances; each non-meeting artifact spec is anchored to its source project and (when applicable) to the meeting it prepares for or follows up on.

As each artifact spec is emitted, it also declares its cross-artifact references (e.g., a prep doc that points at an upcoming meeting, or an email thread that follows up on a decision); these references become edges in an explicit \textit{dependency graph}, and we use Kahn's algorithm with calendar-timestamp tiebreaking to obtain a topological generation order. Each artifact reads a scoped snapshot of the workspace built from its transitive dependencies' events (termed \textsc{WorkspaceEvent}s) and emits new such events as it runs, so later artifacts (including later meetings) inherit a consistent history of decisions, opened and closed action items, and project status changes. A reducer function consolidates these events in timestamp order (with deterministic tiebreaking) to produce the final \textsc{WorkspaceState}. Because the graph makes every conditional dependency explicit, artifacts that are mutually independent under it can be generated in parallel.

\paragraph{4) Turns.} The fourth level generates turn-by-turn dialogue conditioned on personas, agenda, and the latent state. For each meeting, the initial transcript first passes through the adversarial refinement loop (\S\ref{sec:dataset:refinement}); only the post-refinement turns are used to extract that meeting's \textsc{WorkspaceEvent}s, so any structural rewrites the refiner accepts propagate into the final \textsc{WorkspaceState}. We render three transcript variants from the same turn sequence: an initial \emph{canonical} written form used for downstream references, a \emph{semantic-verbatim} form with disfluencies and repairs, and a \emph{TTS} form with non-verbal tokens for speech synthesis.

A final post-processing pass derives reference summary artifacts from the canonical transcript plus latent state, yielding gold references for the summarization metrics in \S\ref{sec:experiments:metrics}.

\subsection{Multimodal rendering}
\label{sec:dataset:rendering}

The TTS form of each transcript is rendered into multi-speaker audio with the Qwen3-TTS 1.7B base model and 12Hz tokenizer \citep{qwen3-tts-2025}. For voice cloning, we draw persona-matched spontaneous-speech reference clips from CommonVoice \citep{ardila-etal-2020-common-voice} (Appendix~\ref{sec:appendix:voice-assignment} details the persona-to-voice matching) and use each clip once to synthesize a short per-speaker enrollment utterance; this enrollment is then frozen and reused as the voice-cloning prompt for every subsequent turn from that speaker. Notably, we perform significant filtering and preprocessing of voice prompts to reduce degenerate TTS outputs and use time and noise-based filtering to perform rejection sampling on final audio generations.

Turns are synthesized independently, mixed with realistic gaps and limited overlap, and post-processed with room simulation. Additional rendering, alignment, acoustic, and multilingual details appear in Appendices~\ref{sec:appendix:tts-rendering}--\ref{sec:appendix:multilingual}.

\subsection{Corpus statistics and release}
\label{sec:dataset:stats}

The released corpus is English-only and comprises 160 workspaces, 2{,}126 meetings, and 570 hours of synthesized audio, spanning domains that include enterprise software development, sales pipelines, customer support, internal operations, medical, academic and non-business scenarios. Appendix~\ref{sec:appendix:corpus-stats} gives aggregate corpus statistics and per-workspace artifact counts, and Appendix~\ref{sec:appendix:workspace-walkthrough} walks through one full workspace end-to-end with a representative instance of every released artifact type; Table~\ref{tab:benchmark-comparison} situates \textsc{OSynth} alongside existing meeting corpora along the axes of scale, available context, and the naturalness of the underlying recordings.

Meeting length averages 16.1 minutes of active speech ($\sim$2{,}900 spoken words per meeting at $\sim$180\,wpm). Unlike popular meeting corpora (AMI, ICSI, MeetingBank), whose transcripts may have entered the pretraining data of modern LLMs, \textsc{OSynth} was newly generated for this work. Also, the pipeline can be re-invoked at low marginal cost to produce additional workspaces, so fresh held-out splits can be created as evaluator models continue to evolve.

\subsection{Adversarial Transcript Refinement}
\label{sec:dataset:refinement}

LLM-generated transcripts possess a characteristic set of structural features that distinguishes them from real transcripts such as clean narrative arcs, perfect turn taking, and few dropped threads. To make the canonical transcripts realistic, we apply an adversarial structural realism pass (\S\ref{sec:dataset:rendering}). For this, we use AMI \citep{carletta-etal-2005-ami} excerpts as style references, and a multi-LLM-family loop pairing a \texttt{gemini-3.1-pro-preview} discriminator with a \texttt{claude-opus-4-7} refiner that revises the transcript for up to five iterations. The discriminator is presented with a pairwise A/B comparison between the candidate and a length-matched AMI excerpt drawn from a held-out AMI split, with speaker labels swapped with generic identifiers (\emph{Speaker 1}, \emph{Speaker 2},~\ldots), so that it cannot anchor on transcript length or speaker names; this adapts the self-refinement structure of \citet{madaan-etal-2023-self-refine} by replacing self-feedback with a cross-family adversarial judge. We split the loop across model families because in preliminary runs a same-family judge was fooled after a single iteration even when obvious structural tells remained; and we cap iterations in the interest of runtime rather than convergence, since test runs with $N=10$ continued to land accepted edits past iteration five. Appendix~\ref{sec:appendix:refinement} gives the full procedure.

\subsection{Realism: human discrimination study}
\label{sec:dataset:realism}

To verify that \textsc{OSynth} is a credible proxy for real workspace meetings, we run a two-alternative forced-choice (2AFC) human-discrimination study. This approach was previously used to assess the realism of synthetic speech \citep{mai-etal-2023-warning}. Raters are shown one real excerpt (from AMI \citep{carletta-etal-2005-ami} or NOTSOFAR-1 \citep{vinnikov-etal-2024-notsofar}) and one synthetic excerpt (from either \textsc{OSynth} or a baseline Haiku w/o obfuscation control) and asked to identify the synthetic side. We retain only sessions that pass two attention checks; Appendix~\ref{sec:appendix:annotation-ui} describes the annotation interface, and Appendix~\ref{sec:appendix:realism-details} the protocol details together with a complementary full-transcript LLM-judge probe.

Among the final set of annotators (24 annotators, 432 non-attention trials), Haiku-w/o-obfuscation fakes were identified reliably: 43/54 (79.6\%) were classified as LLM-generated. \textsc{OSynth} fakes were substantially harder: 213/378 (56.3\%) were classified correctly. The gap is large (+23.3\,pp; Fisher two-sided $p=0.001$), supporting the interpretation that annotators can catch obvious synthetic artifacts while realistic \textsc{OSynth} transcripts lack comparably obvious tells. We note that 56.3\% is above chance: a naive binomial test against 50\% gives $p=0.016$, so \textsc{OSynth} excerpts are still detectable with above chance accuracy rather than indistinguishable. Table~\ref{tab:realism} reports annotator discrimination accuracy.

\begin{table}[t]
  \centering
  \small
  \setlength{\tabcolsep}{6pt}
  \begin{tabular}{lrr}
    \toprule
    \textbf{Synthetic source} & \textbf{Trials} & \textbf{Accuracy} \\
    \midrule
    Haiku (w/o obfuscation) & 54 & 79.6\% \\
    \textsc{OSynth} & 378 & 56.3\% \\
    \midrule
    Chance baseline & --- & 50\% \\
    \bottomrule
  \end{tabular}
  \caption{Human discrimination accuracy in the 2AFC realism study. The \textsc{OSynth} condition is relatively closer to chance, supporting \textsc{OSynth} as a more credible proxy for real workspace meetings.}
  \label{tab:realism}
\end{table}

\section{Context-sensitive evaluation}
\label{sec:experiments}

\begin{table*}[!t]
  \centering
  \footnotesize
  \setlength{\tabcolsep}{4pt}
  \begin{tabular}{llrrrrrr}
    \toprule
    \textbf{System} & \textbf{Ctx.} & \textbf{WER}~$\downarrow$ & \textbf{Num.\ rec.}~$\uparrow$ & \textbf{Term rec.}~$\uparrow$ & \textbf{AI desc.}~$\uparrow$ & \textbf{Assignee}~$\uparrow$ & \textbf{Spk.\ cons.}~$\uparrow$ \\
    \midrule
    \multirow{3}{*}{OpenAI}
      & L1 & 0.162\pmci{.014} & 0.680\pmci{.056} & 0.684\pmci{.040} & 3.49\pmci{.06} & 0.485\pmci{.040} & 0.220\pmci{.045} \\
      & L2 & 0.155\pmci{.015} & 0.695\pmci{.056} & 0.708\pmci{.040} & 3.60\pmci{.06} & 0.504\pmci{.040} & 0.219\pmci{.048} \\
      & L3 & \textbf{0.144}\pmci{.015} & \textbf{0.883}\pmci{.038} & \textbf{0.778}\pmci{.036} & \textbf{3.96}\pmci{.05} & \textbf{0.690}\pmci{.035} & \textbf{0.476}\pmci{.067} \\
    \midrule
    \multirow{3}{*}{Gemini}
      & L1 & 0.099\pmci{.015} & 0.960\pmci{.022} & 0.809\pmci{.031} & 3.55\pmci{.05} & 0.069\pmci{.024} & 0.265\pmci{.050} \\
      & L2 & 0.099\pmci{.015} & 0.963\pmci{.021} & 0.796\pmci{.032} & 3.60\pmci{.05} & 0.071\pmci{.023} & 0.254\pmci{.049} \\
      & L3 & \textbf{0.095}\pmci{.015} & \textbf{0.975}\pmci{.017} & \textbf{0.841}\pmci{.028} & \textbf{3.80}\pmci{.06} & \textbf{0.630}\pmci{.040} & \textbf{0.349}\pmci{.068} \\
    \bottomrule
  \end{tabular}
  \caption{Main results across context levels L1 to L3 on the evaluation set. Values are means with 95\% bootstrap percentile interval (1{,}000 resamples). Bold marks the best context level for each system. Both systems improve mostly monotonically on the entity-sensitive dimensions (number recall, technical-term recall, assignee accuracy, speaker consistency). The OpenAI stack uses \texttt{gpt-4o-transcribe-diarize}, \texttt{gpt-audio}, and \texttt{gpt-5.5}; the Gemini stack uses \texttt{gemini-3-flash} and \texttt{gemini-3.1-pro-preview}.}
  \label{tab:headline}
\end{table*}

\begin{table}[!t]
  \centering
  \scriptsize
  \setlength{\tabcolsep}{4pt}
  \begin{tabular}{llrrr}
    \toprule
    \textbf{Metric} & \textbf{Type} & \textbf{L1} & \textbf{L3} & \textbf{$\Delta$} \\
    \midrule
    Number recall          & iso. & 0.680\pmci{.056} & 0.883\pmci{.038} & +0.20 \\
    Technical-term rec.    & iso. & 0.684\pmci{.040} & 0.778\pmci{.036} & +0.09 \\
    AI assignee accuracy   & iso. & 0.485\pmci{.040} & 0.690\pmci{.035} & +0.20 \\
    Spk.-consistency rate  & iso. & 0.220\pmci{.045} & 0.476\pmci{.067} & +0.26 \\
    \midrule
    LLM-judge faithfulness & std. & 3.10\pmci{.09}  & 3.40\pmci{.10}  & +0.30 \\
    LLM-judge coherence    & std. & 4.60\pmci{.06}  & 4.91\pmci{.04}  & +0.31 \\
    AI description quality & std. & 3.49\pmci{.06}  & 3.96\pmci{.05}  & +0.47 \\
    \bottomrule
  \end{tabular}
  \caption{L1 to L3 gain on isomorphic latent-state metrics versus standard LLM-as-judge summarization metrics for the OpenAI system on the evaluation set including 95\% bootstrap percentile intervals (1{,}000 resamples). The \textbf{Type} column marks each row as either \emph{iso.}\ (isomorphic, scored by exact comparison against \textsc{OSynth}'s latent state) or \emph{std.}\ (standard meeting-summarization metric, scored by LLM-as-judge over the produced artifact). Gains concentrate on the entity-sensitive isomorphic dimensions (assignee accuracy, number recall, speaker consistency), which standard summary-side LLM-as-judge scores systematically compress.}
  \label{tab:isomorphic-vs-standard}
\end{table}

We use \textsc{OSynth} to evaluate two state-of-the-art, multimodal large language model (MLLM) families as meeting understanding systems (\S\ref{sec:experiments:systems}) across three progressively richer levels of workspace context (\S\ref{sec:experiments:context}). Metrics cover transcription, diarization and speaker identification, and downstream summarization artifacts (\S\ref{sec:experiments:metrics}), split into \emph{isomorphic} metrics scored against \textsc{OSynth}'s latent state and \emph{standard} meeting-summarization metrics. We therefore treat meeting summarization as the concrete output surface through which broader meeting understanding can be evaluated. All numbers reported in \S\ref{sec:results} are computed on a held-out evaluation set of 238 meetings drawn from a sample of 15 workspaces; each meeting is scored in each of the three context settings for each system. Point estimates are reported with 95\% bootstrapped confidence intervals (1{,}000 resamples).

\subsection{Systems evaluated}
\label{sec:experiments:systems}

These systems are chosen to reflect the current state of the art in transcription, diarization, speaker identification, and summarization. For reproducibility, Appendix~\ref{sec:appendix:speechbrain} additionally reports a fully open-source SpeechBrain-based cascade on a subset of the same evaluation set.

\paragraph{OpenAI.} A single-vendor cascade built from OpenAI's current API surface: \texttt{gpt-4o-transcribe-diarize} \citep{openai-2026-gpt-4o-transcribe-diarize} produces speaker-attributed transcripts, \texttt{gpt-audio} \citep{openai-2026-gpt-audio} performs enrollment-based speaker identification, and \texttt{gpt-5.5} \citep{openai-2026-gpt-5-5} performs jargon post-correction plus prompted summary, action-item, and outline generation.

\paragraph{Gemini.} A stage-for-stage mirror of the OpenAI pipeline via Vertex AI. Diarized transcription, jargon post-correction, and prompted summary / action-item / outline generation are performed by \texttt{gemini-3-flash} \citep{google-2025-gemini-3-flash}, and enrollment-based speaker identification is performed by \texttt{gemini-3.1-pro-preview} \citep{google-2026-gemini-3-pro}, which we use for the SID stage because of its stronger audio-grounded reasoning on short enrollment clips.

\subsection{Context levels}
\label{sec:experiments:context}

We define three context levels, constructed from \textsc{OSynth}'s latent state, that progressively expand the workspace signal available to a system. Each level is a strict superset of the previous one, so improvement curves across L1--L3 are interpretable as marginal gains from additional context rather than from a change of input modality.

\paragraph{L1 --- No context.} The system sees only the meeting as raw audio. This is the standard meeting-summarization setting reproduced from prior benchmarks \citep{carletta-etal-2005-ami, janin-etal-2003-icsi, hu-etal-2023-meetingbank}.

\paragraph{L2 --- Prior-meeting outputs.} L1 plus the system's own outputs (transcript, summary, action items, outline) for prior meetings in the same workspace, presented as in-context history. L2 isolates the contribution of approximated, longitudinal meeting context without introducing any external workspace signal.

\paragraph{L3 --- Workspace artifacts.} L2 plus the workspace artifacts that surround a meeting: per-speaker voice enrollments and name lists, slide decks tied to the meeting agenda, and inter-meeting emails referencing relevant projects and people. L3 is the deployed-system setting that motivates this work, and the level at which a context-aware system should outperform a context-free one.

Context is encoded as additional prompt fields or model inputs depending on the system architecture; speaker-identity voice enrollments are routed to the speaker-identification stage when a system supports them.

\subsection{Metrics}
\label{sec:experiments:metrics}

We compute metrics in four families: transcription quality, contextualized transcription, diarization and speaker identification, and downstream summarization artifacts. LLM-as-judge metrics use a fixed judge model (\texttt{claude-sonnet-4-6} \citep{anthropic-2026-claude-sonnet}). Because this is the same model family used to generate the corpus, we repeat the judged metrics with an open-weights judge in Appendix~\ref{sec:appendix:qwen-judge}; the L1$\to$L3 improvement is preserved, while the absolute range and calibration of the scores is judge-dependent. We split metrics into \emph{isomorphic} metrics, which are deterministic functions of \textsc{OSynth}'s latent state, and \emph{standard} reference-free LLM-as-judge metrics. \textsc{OSynth}'s latent state makes it possible to score entity-sensitive ASR and summarization behavior that existing meeting benchmarks cannot expose: attendee names and nicknames, project and product names, numbers and dates, technical terms, named-speaker attribution, action-item assignees, and longitudinal entity tracking.

We define each metric reported in Tables~\ref{tab:headline} and~\ref{tab:isomorphic-vs-standard} below; Appendix~\ref{sec:appendix:eval-details} gives the corresponding scoring details. \textbf{Word error rate} (WER) is computed against the semantic-verbatim transcript (\S\ref{sec:dataset:rendering}), which matches the synthesized audio rather than the cleaner canonical written form; both reference and hypothesis are passed through Whisper's \texttt{EnglishTextNormalizer} \citep{radford-etal-2023-whisper} before scoring. \textbf{Number recall} and \textbf{technical-term recall} (iso.) score the fraction of oracle-transcript numbers and dates, and of workspace-jargon terms (acronyms, product names, domain-specific terminology), recovered in the hypothesis; reference term sets are extracted directly from the latent state and intersected with the oracle transcript. \textbf{Cross-meeting speaker-identity consistency} (iso.) is the rate at which a workspace persona is mapped to the same predicted name across meetings within the workspace, exposing labeling-stability failures that a per-meeting attribution score misses. \textbf{Action-item assignee accuracy} (iso.) scores whether each predicted action item is assigned to the correct persona, under partial-credit substring matching against the canonical assignee name for OpenAI (Table~\ref{tab:isomorphic-vs-standard}) and exact match for Gemini (Appendix~\ref{sec:appendix:gemini-iso-std}). \textbf{Action-item description quality} (std.; ``AI desc.'' in Table~\ref{tab:headline}) is an LLM-judge 1--5 rating of how completely and concretely each predicted action-item description captures the underlying commitment, scored against the gold action-item list derived from the latent state. \textbf{LLM-judge faithfulness} and \textbf{coherence} (std.) score per-meeting summaries on the corresponding 1--5 dimensions following \citet{liu-etal-2023-g-eval} and \citet{zheng-etal-2023-judging} against the gold summary; see Appendix~\ref{sec:appendix:prompts} for details.

\section{Results}
\label{sec:results}

We report per-system results across the three context levels L1--L3 on the metric families defined in \S\ref{sec:experiments:metrics}, computed on the 238-meeting evaluation set. Table~\ref{tab:headline} summarizes the main results; Table~\ref{tab:isomorphic-vs-standard} compares the context gains visible through isomorphic metrics and standard LLM-as-judge metrics. The open-source SpeechBrain cascade is reported separately in Appendix~\ref{sec:appendix:speechbrain}. Two robustness checks accompany these results: the judged metrics recomputed with an open-weights judge (Appendix~\ref{sec:appendix:qwen-judge}), and an anchoring study on real AMI meetings (\S\ref{sec:anchor}).

\subsection{Trends across context levels}
\label{sec:results:headline}

For the OpenAI system, summary-side artifact quality and contextualized-transcription recall improve monotonically from L1 to L3 (Table~\ref{tab:headline}). Gains concentrate on the contextualized-transcription and entity-sensitive summarization dimensions: number recall (+0.20 absolute, L1$\to$L3), technical-term recall (+0.09), action-item description quality (+0.47 points on a 1--5 scale), action-item assignee accuracy (+0.20, scored under partial-credit substring matching), and cross-meeting speaker-identity consistency (+0.26). Concretely, the +0.20 number-recall jump is driven by recoveries such as L1's ``\emph{PAX eight}'' for the ISO~20022 identifier ``\texttt{pacs.008}'' (anchored at L3 by the workspace project document); a side-by-side excerpt is given in Appendix~\ref{sec:appendix:error-examples}.
The Gemini stack shows the same broad pattern (Table~\ref{tab:headline}), with the largest gains again on entity-sensitive dimensions: assignee accuracy, number recall, and technical-term recall all rise from L1 to L3. The assignee jump is expected because L3 adds per-speaker voice enrollments, turning assignee recovery from an open-ended name-generation problem (where names must be inferred from contextual mentions) into a closed-set lookup against workspace identities. On the OpenAI side, the +0.20 assignee gain is concretely illustrated by the L1$\to$L3 trajectory \emph{Bevan~$\to$~Devin~$\to$~Devon Ashby} (an invented name at L1, a different invented name at L2, the canonical workspace identity only at L3); the full block is in Appendix~\ref{sec:appendix:error-examples}.
Overall WER improves only modestly with context (OpenAI L1$\to$L3: $0.162\to0.144$; Gemini: $0.099\to0.095$), and in both cases the bootstrap 95\% confidence intervals for L1 and L3 overlap. The entity-sensitive dimensions move several times as far over the same increment, consistent with residual errors routing primarily through entity-level surface forms rather than general word recognition, which is precisely where L3 yields the largest gains.

\subsection{Isomorphic versus standard metrics}
\label{sec:results:metrics}

Isomorphic latent-state metrics and standard summarization metrics tell partially different stories (Table~\ref{tab:isomorphic-vs-standard}; Gemini breakdown, on the earlier 50-meeting subset, in Appendix~\ref{sec:appendix:gemini-iso-std}). Both metric families agree that L3 improves over L1 on the main entity-sensitive dimensions, but LLM-as-judge summary-side scores compress the gains visible in the isomorphic metrics. The Gemini stack has one non-monotonic standard cell: LLM-judge faithfulness dips slightly at L3 despite uniform improvement on every isomorphic dimension, reflecting the known tendency of reference-free protocols to underweight factual precision and over-penalize informative workspace-canonical expansions \citep{tang-etal-2024-tofueval}.

\subsection{Which workspace artifacts matter?}
\label{sec:results:ablations}

An OpenAI-stack ablation separates the L3 increment into per-speaker voice \emph{enrollments} and \emph{workspace text} artifacts (member lists, emails, slides, and agendas). Workspace text explains almost all of the number-recall gain, while assignee accuracy benefits from both streams: enrollments anchor diarization clusters to canonical identities, and workspace text supplies the names and entities the summarizer writes into action items. Appendix~\ref{sec:appendix:ablations} gives the full table and discussion.

Appendix~\ref{sec:errors} further analyzes OpenAI L1 errors using a six-way ASR taxonomy. The dominant categories are named entities, numbers/dates, and technical terms; L3 mainly helps the entity-sensitive cases, while filler, function-word, and lexical reanalysis errors remain largely orthogonal to workspace context.

\section{Anchoring on real meetings}
\label{sec:anchor}

\textsc{OSynth}'s isomorphic metrics require a ground-truth latent state, which real corpora do not provide, so they cannot be transferred to recorded meetings. The judged summary metrics can be, which lets us check the direction of the context effect on real data. We randomly selected two meeting series from the ES subset of AMI \citep{carletta-etal-2005-ami}, ES2002 and ES2003, of four meetings each, and constructed a corresponding workspace state by aggregating the AMI metadata (an anonymous identity map held across each series), the gold transcripts, and the human-written abstractive summaries. We ran the Gemini system in the L1 and L3 settings, using for L3 the system's own outputs on prior meetings in the series together with the meeting title, and scored the resulting summaries with the judged summary metrics (Table~\ref{tab:ami-anchor}). Faithfulness improves by $+0.13$; coherence is at the ceiling in both settings. The direction agrees with the main result, which we take as convergent validity for the context effect rather than as a second measurement of its size.

A major caveat of this study is the limited size, but it further supports the validity of our results by providing convergent evidence on real, recorded audio.

\begin{table}[!htbp]
  \centering
  \footnotesize
  \setlength{\tabcolsep}{4pt}
  \begin{tabular}{lrrr}
    \toprule
    \textbf{Metric} & \textbf{L1} & \textbf{L3} & \textbf{$\Delta$} \\
    \midrule
    LLM-judge faithfulness & 3.50 & 3.63 & +0.13 \\
    LLM-judge coherence    & 5.00 & 5.00 & \phantom{+}0.00 \\
    \bottomrule
  \end{tabular}
  \caption{Anchoring study on eight real AMI meetings (series ES2002 and ES2003), Gemini system, judged summary metrics only.}
  \label{tab:ami-anchor}
\end{table}

\section{Discussion}
\label{sec:discussion}

Both state-of-the-art systems show monotonic improvement on the entity-sensitive isomorphic dimensions as workspace context is added, with the largest gains on action-item assignee accuracy, number recall, technical-term recall, and cross-meeting speaker-identity consistency. Notably, these are all dimensions that prior meeting benchmarks cannot easily expose due to limited context, anonymization of named entities, and lack of the extensive oracle latents we use for evaluation. These gains do not mostly come from better acoustic recognition. They come from workspace-specific evidence: voice enrollments map diarization clusters to people, while workspace names and inter-meeting emails help resolve the entities that should appear in summaries. As a result, benchmarks that evaluate only isolated meetings, for instance, with no workspace context, understate the quality that a deployed workspace-aware system can achieve. The ablation in \S\ref{sec:results:ablations} further shows that the two L3 streams (enrollments and workspace text) are complementary rather than redundant.

This pattern bears directly on a long-running critique that WER ``does not reflect text understanding and only weakly correlates with human judgement of a transcript's quality'' \citep[as summarized by][]{kuhn-etal-2023-asr-accuracy}, originally argued by \citet{wang-etal-2003-wer-indicator}, \citet{mishra-etal-2011-perceived-accuracy}, and \citet{favre-etal-2013-asr-utility}. By weighting every token equally, WER absorbs context-driven gains into the long tail of low-value words, even when, as in our setting, those gains concentrate on the entities that determine the usefulness of a transcript or downstream summary. The isomorphic, entity-grounded metrics enabled by \textsc{OSynth}'s latent state expose precisely the dimensions WER does not expose. Thus, the L1 to L3 measurement scales from a modest, aggregate WER improvement into a substantially larger, quantifiable improvement on the other measurements that we have considered.

\section{Conclusion}

We present \textsc{OSynth}, a longitudinal synthetic meeting corpus with aligned workspace artifacts and an exact latent-state reference, and use it to evaluate state-of-the-art meeting notetaking systems across progressively richer context levels. Human discrimination results support \textsc{OSynth} as a relatively credible proxy for real meeting transcripts as compared to naive LLM generation, and the evaluation shows that workspace context improves summary artifacts and contextualized transcription on entity-sensitive dimensions that prior benchmarks cannot equivalently expose.

\section*{Limitations}

\textsc{OSynth} is a fully synthetic corpus and therefore cannot fully reproduce the long-tail acoustic, sociolinguistic, and interactional phenomena present in recorded workspace meetings; we make no claim that results reported here transfer numerically to natural recordings. We make this choice deliberately: a fully synthetic pipeline is what unlocks the longitudinal, multi-artifact workspace setting at releasable scale and with an exact latent-state reference. Scaling our AMI grounding study to more convergent results on natural spoken meeting data is a worthwhile point for future investigation.

Additionally, the released TTS audio is rendered with a single multi-speaker model and a single voice-cloning prompt per persona; even when the cloning prompt is drawn from spontaneous conversational speech, the resulting per-turn renderings can still exhibit a perceptibly read-aloud quality (i.e., steady prosody) relative to fully spontaneous speech, and downstream ASR and diarization numbers should therefore be read as relative comparisons within \textsc{OSynth}, not as absolute estimates of performance on natural recordings.

The reference clips themselves are drawn from the Mozilla Common Voice \emph{Spontaneous Speech} corpus rather than the scripted releases, so the read-aloud quality is introduced downstream: measured against the cloning prompts, \textsc{OSynth} turns are paced more uniformly (Appendix~\ref{sec:appendix:pacing}). A second limitation of the artifacts is that they are clean by construction: workspace emails, slides, and documents contain no typos or transcription noise beyond the ASR output itself, and injecting such noise is left to future work.

Finally, the evaluation in \S\ref{sec:experiments} fixes a single prompt template per task and a single decoding configuration per system; absolute scores will shift under heavier prompt engineering. We furthermore ensure diversity only via the workspace random initialization which would become a limitation as dataset size is scaled up.

\section*{Ethical considerations}

\textsc{OSynth} is generated end-to-end by an LLM-based pipeline. All personas, organizations, project names, and email addresses are sampled from generated pools rather than drawn from real entities, and TTS voices are cloned from a curated subset of CommonVoice \citep{ardila-etal-2020-common-voice} whose speakers have consented to broad reuse. Because these names are generated without reference to real people or companies, any resemblance a generated name bears to an actual person or organization is coincidental and unintended; references to public standards and to the bodies that maintain them appear only as background domain context.

That being said, the CommonVoice terms of use prohibit any attempt to re-identify the speakers behind the released voice clips, and we make no such attempt at any stage of the \textsc{OSynth} pipeline: cloning prompts are used solely as anonymous acoustic conditioning for the TTS model, are stripped of all CommonVoice metadata before use, and are paired with synthetic personas that bear no relation to the underlying contributor. Downstream users of the released audio inherit the same condition and should not attempt to link voices to real individuals.

Because the corpus is synthetic, it may encode biases of the generating LLM in persona attributes, dialogue style, and decision dynamics. The corpus is intended for evaluation of meeting-summarization and notetaking systems; it should not be used as training data for systems whose claims of generalization rest on naturalistic conversational behavior.

Finally, an LLM-based coding assistant was used to copy-edit and copy-check the final draft of this manuscript.

\section*{Acknowledgments}

We thank Sam Liang and Yun Fu for their support of this work, the annotators who contributed to the human-discrimination study, and the anonymous reviewers for their constructive feedback.


\bibliography{custom}

\appendix

\section{Corpus statistics}
\label{sec:appendix:corpus-stats}

\begin{table}[h]
  \centering
  \small
  \setlength{\tabcolsep}{6pt}
  \begin{tabular}{lr}
    \toprule
    Workspaces & 160 \\
    Meetings & 2{,}126 \\
    Hours of audio & 570 \\
    \midrule
    Meetings / workspace (mean) & 13.3 \\
    Personas / workspace (mean) & 12.9 \\
    Email threads / workspace (mean) & 4.3 \\
    Shared documents / workspace (mean) & 1.5 \\
    \bottomrule
  \end{tabular}
  \caption{\textsc{OSynth} corpus statistics. Each workspace bundles meeting audio, aligned transcripts, downstream notetaking artifacts, inter-meeting workspace artifacts, and per-speaker enrollments derived from a shared latent state. Workspaces are generated with varying workspace size and time horizons, thus subsamples of the data follow distinct distributions in terms of length and content.}
  \label{tab:corpus-stats}
\end{table}

\section{Workspace walkthrough example}
\label{sec:appendix:workspace-walkthrough}

To give a concrete picture of how a single \textsc{OSynth} workspace appears across personas, projects, meetings, and artifacts, we walk through one representative workspace: a synthetic Series-C fintech midmarket organization. The organization is observed over a two-week window.

\paragraph{Org chart and personas.} The active org chart contains ten speaking personas plus two external contacts (Charlotte Bewick of Brindlewood; Tomás Herrera of Nexbridge Capital). Reporting lines are: Renata Sokolowski (CEO, \texttt{p1}) over Marcus Delacroix (VP Eng, \texttt{p2}; reports: Simone Okafor, Rafael Quintero, Devon Ashby), Priya Nambiar (Dir. Product, \texttt{p3}; report: Alejandro Vásquez), Dillon Hargrove (VP Sales, \texttt{p4}; reports: Adriana Castellanos, Everett Cho), and Yuki Tanaka (Dir. Ops/Compliance, \texttt{p5}; report: Devansh Pillai). Two personas (\emph{Marcus}, \emph{Dillon}) carry an explicit nickname policy (\emph{Marc}, \emph{Dill}; Appendix~\ref{sec:appendix:nicknames}), which the transcript LLM uses to vary how other speakers refer to them.

\paragraph{Active projects.} Three projects are simultaneously live in this window: (i) \emph{ISO 20022 Payment Message Migration} (owner: Yuki Tanaka; blocked on Payments Canada draft fields; target 2026-07-31), (ii) \emph{Clearflow Orchestrate Failover Hardening} (owner: Simone Okafor; blocked on locking strategy; target 2026-06-19), and (iii) \emph{Enterprise Prospect Acceleration --- Brindlewood Commerce Group} (owner: Dillon Hargrove; SOC~2 and BDW-sandbox prerequisites; target 2026-05-31). Each project carries its own document set, calendar entries, and \textsc{WorkspaceEvent} stream (\S\ref{sec:dataset:pipeline}).

\paragraph{Meeting calendar.} The workspace contains 28 meetings across the two-week window, spanning weekly executive syncs, daily standups, recurring 1:1s, and ad-hoc design and project reviews. Table~\ref{tab:walkthrough-calendar} shows a representative slice of the first three days centered on the focus meeting used in the artifact excerpts below.

\begin{table}[!t]
  \centering
  \footnotesize
  \setlength{\tabcolsep}{3pt}
  \begin{tabular}{@{}l p{4.2cm} r r@{}}
    \toprule
    \textbf{Date} & \textbf{Title} & \textbf{Dur} & \textbf{Att} \\
    \midrule
    05-18 09:00 & Executive Leadership Sync                & 60 & 5 \\
    05-18 09:15 & Engineering Daily Standup                & 15 & 4 \\
    05-18 11:00 & Project Brindle Kickoff                  & 45 & 4 \\
    05-19 09:15 & Engineering Daily Standup                & 15 & 4 \\
    05-19 10:00 & \textbf{Marcus--Simone 1:1} ($\star$)    & 30 & 2 \\
    05-19 10:00 & ISO 20022 Migration Sync                 & 60 & 5 \\
    05-19 10:30 & Marcus--Rafael 1:1                       & 30 & 2 \\
    05-19 11:00 & Marcus--Devon 1:1                        & 30 & 2 \\
    05-20 09:00 & SOC 2 Package Prep Session               & 60 & 2 \\
    05-20 14:00 & Orchestrate Failover Review              & 60 & 5 \\
    05-20 15:00 & CAS Locking Strategy Review              & 90 & 3 \\
    05-21 10:00 & BDW Sandbox Scope (Devon \& Adriana)     & 30 & 2 \\
    \bottomrule
  \end{tabular}
  \caption{Three-day slice of the example workspace meeting calendar around the focus meeting (marked with $\star$). \emph{Dur} is scheduled duration in minutes; \emph{Att} is attendee count. The full workspace contains 28 meetings across the two-week observation window.}
  \label{tab:walkthrough-calendar}
\end{table}

\paragraph{Representative artifacts.} The focus meeting (\emph{Marcus $\leftrightarrow$ Simone 1:1}, 2026-05-19 10:00) is a 30-minute recurring 1:1 between the VP of Engineering and a senior engineer. \\

\emph{(1) Transcript excerpt (canonical, 6 turns).}

\begin{quote}\small
\textbf{Marcus:} Hey. Failover. \\
\textbf{Simone:} Race condition's isolated. It's the locking now --- registry, concurrent writes. \\
\textbf{Marcus:} Right. \\
\textbf{Simone:} Global lock's out, P99 was like 120. \\
\textbf{Simone:} Devon's on CAS. The harness doesn't replay reconnects, so we don't actually know what happens to in-flight transitions when a node drops. \\
\textbf{Marcus:} When's he have something reviewable? Not merged. \\
\textbf{Simone:} End of week realistically. Brindlewood's pulling on him.
\end{quote}

\emph{(2) Gold summary (1 paragraph).}

\begin{quote}\small
Marcus and Simone covered the status of the Orchestrate Failover Hardening project, which remains blocked pending Devon's lock-free CAS prototype review; Simone committed to syncing with Devon before end of week and flagging Marcus by Thursday if the approach doesn't look viable. They briefly touched on the ISO 20022 SettleMap work, which is in a holding pattern on three \texttt{camt.054} sub-fields awaiting Payments Canada guidance, and discussed Devon's split availability between the CAS prototype and the Brindlewood sandbox environment.
\end{quote}

\emph{(3) Gold action items (selected).}

\begin{quote}\small
\textbf{[Marcus Delacroix]} Handle Devon's competing priorities with Brindlewood to keep him focused on CAS work (due before EOW). \\
\textbf{[Simone Okafor]} Look into whether SettleMap mappings that don't touch \texttt{camt.054} can be sequenced and progressed independently despite XSD cross-references. \\
\textbf{[Simone Okafor]} Check the Thursday meeting invite and confirm whether Karol moved it to 2 or 2:30, then tell Hwang whether attending remote or in-room.
\end{quote}

\emph{(4) Calendar invite (subject and body).}

\begin{quote}\small
\textbf{Subject:} Marcus $\leftrightarrow$ Simone 1:1 \\
\textbf{When:} Tuesday, 2026-05-19, 10:00--10:30 ET (30 min) \\
\textbf{Where:} Conference Room 1 (hybrid: Marcus in-room, Simone remote) \\
\textbf{Body:} Recurring 1:1. Touch base on Orchestrate failover hardening status, ISO 20022 SettleMap progress, and Devon's availability on Brindlewood / CAS work.
\end{quote}

\emph{(5) Email thread (1-message excerpt).}

\begin{quote}\small
\textbf{From:} Yuki Tanaka $\langle$\texttt{yuki.tanaka@\allowbreak clearflow\allowbreak payment\allowbreak infrastructure.com}$\rangle$ \\
\textbf{To:} Charlotte Bewick (Brindlewood) \quad \textbf{Cc:} Dillon Hargrove, Devansh Pillai \\
\textbf{Subject:} SOC 2 Type II package delivered to Brindlewood --- FYI + Q\&A call offer \\
``Hi Charlotte, I wanted to confirm that Clearflow's SOC 2 Type II attestation package has been shared with you via our secure document portal. No exceptions were noted that would materially affect availability, security, or processing integrity. Happy to set up a 30-minute Q\&A call with myself and Devansh whenever it's useful. Best, Yuki.''
\end{quote}

\emph{(6) Shared document excerpt (project doc).}

\begin{quote}\small
Clearflow Payments --- SOC 2 Type II Compliance Posture Summary. Prepared for: Brindlewood Commerce Group (Procurement \& Vendor Risk). Audit Period: April 1, 2024 to March 31, 2025. Independent Auditor: Hartwell \& Crane LLP. Trust Service Criteria: Security, Availability, Confidentiality, Processing Integrity. Audit covered Clearflow's core payment infrastructure platform in its entirety (Clearflow Orchestrate, Clearflow Settle, Clearflow Shield). One qualified opinion (CF-CM-04, UK change-management workflow); fully remediated as of 2026-02-03.
\end{quote}

\emph{(7) Enrollment metadata.} Per-persona enrollment clips are short (7--13\,s), drawn from CommonVoice \citep{ardila-etal-2020-common-voice} after the curation pipeline of Appendix~\ref{sec:appendix:tts-rendering}.

\section{Generation-pipeline details}
\label{sec:appendix:generation-details}

The schedule is frontier-parallel: at each step, every artifact whose predecessors have completed is dispatched concurrently against a \emph{scoped} workspace state built by replaying only its transitive-dependency events. Artifacts emit typed \texttt{WorkspaceEvent} payloads (decisions, action items opened or closed, project status changes), which a final reconciliation pass globally sorts by timestamp and folds through a pure reducer \(\mathrm{reduce}:\mathrm{State}\times\mathrm{Event}\to\mathrm{State}\).

Transcript prompts receive a calendar-anchored word budget \((\textit{floor},\textit{target},\textit{ceiling})\) derived from scheduled duration and a target words-per-minute range. Transcripts below the floor are regenerated once with a deficit-aware addendum, and the downstream realism rewrite (\S\ref{sec:dataset:refinement}) is gated against the same floor so that structural realism cannot collapse a transcript below its scheduled-duration contract.

\section{Workspace-artifact ablation details}
\label{sec:appendix:ablations}

To isolate which components of L3 drive improvement over L2, we group the L3 increment into two streams---per-speaker voice \emph{enrollments} (routed to the speaker-identification stage) and \emph{workspace text} artifacts (the workspace member list, inter-meeting emails, slide decks, and meeting agendas)---and run the OpenAI stack with each stream individually enabled on top of L2 (Table~\ref{tab:ablation}).

The three target dimensions show qualitatively different complementarity patterns. \textbf{Number recall} is dominated almost entirely by workspace text: text-only already captures the full L3 gain ($0.637\to0.975$), while enrollments-only barely moves the number above L2 ($0.637\to0.605$). \textbf{Technical-term recall} shows a similar but weaker text-dominated pattern (text-only $0.723\to0.739$ vs.\ full $0.723\to0.760$). \textbf{Action-item assignee accuracy} is the cleanest case of two-stream complementarity: workspace text alone improves from $0.524\to0.752$, enrollments alone from $0.524\to0.698$, and full L3 exceeds either single stream ($0.524\to0.792$). Enrollments anchor SID cluster labels to canonical workspace identities, while workspace text supplies the canonical names, projects, and entities that the summarization stage writes into action items and contextualized terms.

\begin{table}[!t]
  \centering
  \footnotesize
  \setlength{\tabcolsep}{3pt}
  \begin{tabular}{lrrr}
    \toprule
    \textbf{Configuration} & \textbf{Num.\ rec.} & \textbf{Term rec.} & \textbf{Assignee} \\
    \midrule
    L2 (no L3 streams)              & 0.637 & 0.723 & 0.524 \\
    + enrollments only              & 0.605 & 0.657 & 0.698 \\
    + workspace text only           & 0.975 & 0.739 & 0.752 \\
    \textbf{L3 (both streams)}      & \textbf{0.973} & \textbf{0.760} & \textbf{0.792} \\
    \bottomrule
  \end{tabular}
  \caption{OpenAI-stack ablation on the L2$\to$L3 increment, computed on the earlier 50-meeting evaluation subset. Per-speaker voice \emph{enrollments} (routed to SID) and \emph{workspace text} artifacts (member list, emails, slides, agendas) contribute differently across dimensions: workspace text accounts for essentially the entire number-recall gain, while enrollments and workspace text contribute complementarily on assignee accuracy. Number- and term-recall figures use exact matching; assignee accuracy uses partial-credit substring matching.}
  \label{tab:ablation}
\end{table}

\section{Error analysis}
\label{sec:errors}

To understand what drives the headline numbers in \S\ref{sec:results}, we cluster reference--hypothesis disagreements in the OpenAI L1 output and codify a six-way taxonomy of ASR error classes: \textbf{NE} (named-entity confusion), \textbf{NUM} (numeric/date verbalization), \textbf{TERM} (acronym or compound-boundary errors), \textbf{FILL} (disfluency insertion or deletion), \textbf{FUNC} (function-word reshaping), and \textbf{LEX} (common-word bias or local reanalysis).

The three most prevalent categories (\textbf{NE}, \textbf{NUM}, and \textbf{TERM}) together account for roughly two thirds of extracted span pairs and dominate the word-error trends in \S\ref{sec:results:headline}. L3 context mainly helps the entity-sensitive classes: in one representative passage, \textsc{BLE} and \textsc{ArcBus} are recovered only once the workspace jargon list is available, and enrollment-based speaker identification corrects an action-item assignee from a hallucinated name to \emph{Darren Okafor}. NUM errors are less affected because the workspace artifacts contain little numeric metadata, while FILL, FUNC, and LEX are mostly orthogonal to context. Appendix~\ref{sec:appendix:error-examples} gives the full qualitative examples.

\section{Adversarial transcript refinement details}
\label{sec:appendix:refinement}

LLM-generated transcripts possess a characteristic set of structural features that distinguishes them from real transcripts: a clean narrative arc, efficient speaker handoffs, explicit action-item recaps, and near-total absence of dropped threads. The adversarial structural realism pass targets these discourse-level tells only; surface disfluencies (filled pauses, repetitions, truncations) are injected separately by the spoken-form formatter (\S\ref{sec:dataset:rendering}).

A \texttt{claude-sonnet-4-6} rewriter first restructures the transcript using a real AMI dataset \citep{carletta-etal-2005-ami} excerpt as a style reference, breaking the narrative arc (raising and dropping at least one thread, derailing the main topic for several turns, omitting any end-of-meeting recap), adding structurally costly interruptions, replacing expository dialogue with oblique shared-knowledge shorthand, blurring per-speaker roles, and letting the meeting trail off without resolution. The rewriter is allowed to split, merge, or locally reorder turns but is prevented from adding or removing speakers or discarding key decisions.

The rewritten transcript then enters a refinement loop using a cross-family discriminator (\texttt{gemini-3.1-pro-preview}). The discriminator performs a pairwise A/B comparison against an AMI meeting \citep{carletta-etal-2005-ami} excerpt drawn from a holdout split reserved for the \textsc{OSynth} realism loop, with speaker labels replaced by generic identifiers and the candidate trimmed to match the AMI excerpt's line budget inside the judge call; this neutralizes both naming-distribution and length-based shortcut features that would otherwise dominate the verdict regardless of structural realism. The synthetic side is randomized per (meeting, iteration) by a meeting-deterministic RNG so the verdict is reproducible. If the discriminator is fooled, the loop exits early. Otherwise it returns the reasoning for the remaining tells, and a \texttt{claude-opus-4-7} refiner revises the transcript to address them directly. The loop repeats for up to $N=5$ iterations. If the discriminator cites narrative efficiency, the refiner is permitted to permanently delete a topic thread, prioritizing structural realism over content completeness; however, when the candidate is already at or near its calendar-anchored word floor (\S\ref{sec:dataset:pipeline}), this permission is withdrawn for that meeting so that the refinement loop cannot collapse a transcript below its scheduled-duration contract, and the refiner is instructed to restructure rather than shrink.

\paragraph{Why a cross-family loop.} In early development we tried a same-family configuration in which a single Claude model played all three roles (rewriter, refiner, and discriminator). On those runs the discriminator typically declared the candidate indistinguishable on the very first iteration, causing the loop to exit immediately even when the candidate still exhibited the characteristic tells the pass is meant to remove (clean recaps, perfect handoffs, no dropped threads). Switching the discriminator to a different model family from the rewriter/refiner removed this trivial fixed point: the Gemini judge consistently flagged structural tells that the same-family judge accepted, and accepted candidates only after multiple rounds of edits.

\paragraph{Iteration cap.} We chose $N=5$ as a runtime cap rather than as an empirical convergence point. In test runs we extended the loop to $N=10$ and observed that a non-trivial fraction of meetings continued to receive accepted refiner edits past iteration five, so the cap reflects the per-meeting cost budget for the full corpus generation rather than a saturation in the discriminator's verdicts.

\paragraph{Length consideration.} We noticed that our original implementation resulted in adversarial refinement reducing the length of the transcript at each iteration. We updated the prompt and added a minimum length (shorter lengths being rejected), resulting in two possible implementations of adversarial refinement: a naive implementation and a length-preserving implementation.

\section{Additional realism-study details}
\label{sec:appendix:realism-details}

In the 2AFC human-discrimination study (\S\ref{sec:dataset:realism}), annotators were recruited from a pool of professionals working in the meeting-transcription domain. Annotators were informed before beginning that their responses would be used to evaluate synthetic meeting transcripts and reported in aggregate in a research publication. Each session interleaves realistic \textsc{OSynth} trials, Haiku-w/o-obfuscation control trials, and two attention-check trials whose correct answer is unambiguous; only sessions in which both checks were answered correctly are retained for the headline analysis.

As complementary supporting evidence at the meeting level, we run a single-stimulus LLM-judge probe in which \texttt{gpt-5.4} \citep{openai-2026-gpt-5-4} (\texttt{reasoning\_effort=low}) is shown one full meeting transcript at a time and asked to classify it as real or synthetic. On a balanced sample of 100 transcripts (50 real AMI, 50 \textsc{OSynth}), the judge achieves 59\% overall accuracy with 100\% recall on real meetings and only 18\% recall on \textsc{OSynth} meetings, calling 91/100 transcripts ``real'' overall (41 of those being false positives). This probe has a strong ``real'' prior on isolated full transcripts, so we read it as consistent with the 2AFC finding that realistic \textsc{OSynth} excerpts lack obvious tells rather than as independent proof of realism. A qualitative analysis of the free-text justifications raters provided when correctly identifying synthetic excerpts surfaces a small set of recurring cues: over-articulated agendas, near-absence of overlapping speech, and unusually clean topic transitions.

\section{System and metric details}
\label{sec:appendix:eval-details}

Both single-vendor systems use vendor-recommended defaults for decoding and a fixed prompt template per task. For cascaded systems, context is encoded as additional fields in the prompted summarization, action-item, and outline stages; for one-shot MLLMs, context is passed as additional inputs to the single call. Speaker-identity voice enrollments are passed to the speaker-identification stage of cascaded systems and as additional audio inputs to MLLMs that support multi-clip audio inputs; for MLLMs that do not, we pass an ASR transcript of each enrollment together with the speaker's name.

\paragraph{OpenAI.} The \texttt{gpt-4o-transcribe-diarize} endpoint produces a speaker-attributed transcript with cluster labels in a single call. The diarize endpoint does not accept Whisper-style prompt biasing, so workspace vocabulary is instead applied as a downstream post-correction at L3. \texttt{gpt-audio} performs cluster-level speaker identification by comparing a per-cluster representative audio clip against the workspace's speaker enrollments. \texttt{gpt-5.5} performs jargon post-correction together with prompted summarization, action-item, and outline generation, consuming the resulting transcript and the level-conditioned workspace context.

\paragraph{Gemini.} The Gemini baseline mirrors the OpenAI pipeline via Vertex AI but is split across two Gemini models. Stage~1 (\texttt{gemini-3-flash} \citep{google-2025-gemini-3-flash}) ingests the meeting audio directly and emits a JSON array of speaker-tagged segments with cluster labels. Stage~2, enabled at L3, uses \texttt{gemini-3.1-pro-preview} \citep{google-2026-gemini-3-pro} for enrollment-based speaker identification, comparing a per-cluster representative clip against the workspace's enrollment clips and returning the matching display name; we use the larger Pro model for this stage because it is the SID stage that most relies on fine-grained acoustic-comparison reasoning. Stage~3 (\texttt{gemini-3-flash}) consumes the resulting transcript together with the workspace context and produces jargon post-corrections, summaries, action items, and outlines.

\paragraph{Transcription quality.} We report standard word error rate (\textsc{WER}), scored against the semantic-verbatim transcript (\S\ref{sec:dataset:rendering}), with reference and hypothesis passed through Whisper's \texttt{EnglishTextNormalizer} \citep{radford-etal-2023-whisper} before scoring.

\paragraph{Contextualized transcription.} \textsc{OSynth}'s latent state makes it possible to score ASR on word categories that disproportionately affect downstream user experience. We report two such categories: numbers and dates (\textsc{number recall}), and technical and domain-specific terms (\textsc{technical-term recall}). Reference term sets for both are extracted directly from the workspace's latent state and intersected with the oracle transcript, so a term counts against a system only if it was actually spoken in the meeting being scored.

\paragraph{Diarization and speaker identification.} We report cross-meeting speaker-identity consistency: the rate at which a workspace persona is mapped to the same predicted name across the meetings of its workspace. Because the released alignments tie every word to a canonical persona (Appendix~\ref{sec:appendix:alignment}), this scores whether words are routed to the correct \emph{named} speaker rather than merely to a self-consistent per-meeting label, and it exposes labeling-stability failures that a per-meeting score cannot.

\paragraph{Summarization artifacts.} Per-meeting summary quality is scored by an LLM judge on \textsc{faithfulness} and \textsc{coherence} following \citet{liu-etal-2023-g-eval} and \citet{zheng-etal-2023-judging}, with reference content drawn from the gold summary derived from the latent state. Action-item extraction is scored against the latent action-item list by assignee accuracy, together with an LLM judgment of how completely and concretely each predicted description captures the underlying commitment.

\section{Prompt templates}
\label{sec:appendix:prompts}

For reproducibility, we reproduce a representative prompt (or prompt-template sketch) for each major prompting stage of the evaluation pipeline. The summarization, action-item, and outline prompts are shared verbatim across the OpenAI and Gemini stacks (modulo the model handle); the enrollment-based SID and vocabulary post-correction prompts are stack-specific. The adversarial-realism loop (\S\ref{sec:dataset:refinement}) uses two additional model-family-specific prompts, reproduced below. Verbatim prompt text is shown in monospace blocks; templating placeholders are written \texttt{\{name\}}.

\paragraph{Summarization (L1 / L2 / L3).} The OpenAI (\texttt{gpt-5.5}) and Gemini (\texttt{gemini-3-flash}) stacks share a single summarization prompt template (Figure~\ref{fig:prompt-summary}). The user message contains an optional \texttt{CONTEXT} block (empty at L1, populated at L2/L3) followed by a \texttt{TRANSCRIPT} block. At L2 the \texttt{CONTEXT} block contains the system's own prior-meeting outputs (summaries and action items, each truncated); at L3 it additionally contains the workspace name list (canonical name + role per persona), a condensed digest of recent workspace emails, and the meeting's agenda title and type. Each L3 add-on is rendered under a byte budget so the assembled context fits in a single inference call.

\begin{figure*}[!t]
  \centering
  \begin{minipage}{0.97\textwidth}\small\ttfamily
You are generating a meeting summary from a transcript.\\[2pt]
\{context\_block\}TRANSCRIPT:\\
\{transcript\}\\[2pt]
Write a concise summary covering the key discussion points, decisions made, and outcomes. Use bullet points for clarity. Keep it under 500 words.
  \end{minipage}
  \caption{Shared summarization prompt template used by both the OpenAI (\texttt{gpt-5.5}) and Gemini (\texttt{gemini-3-flash}) stacks across all three context levels. \texttt{\{context\_block\}} is empty at L1 and populated at L2/L3 as described in the surrounding prose.}
  \label{fig:prompt-summary}
\end{figure*}

\paragraph{Action-item extraction.} The OpenAI and Gemini stacks share an action-item-extraction prompt template (Figure~\ref{fig:prompt-actions}); the outline prompt has the same shape, with the instruction body replaced by a request for a markdown-heading outline. The same \texttt{CONTEXT} block as in summarization is used.

\begin{figure*}[!t]
  \centering
  \begin{minipage}{0.97\textwidth}\small\ttfamily
You are extracting action items from a meeting transcript.\\[2pt]
\{context\_block\}TRANSCRIPT:\\
\{transcript\}\\[2pt]
Extract all action items mentioned in the meeting. For each, provide:\\
- description: What needs to be done\\
- assignee: Who is responsible (use the speaker name if mentioned, otherwise "Unassigned")\\[2pt]
Respond with ONLY a JSON array of \{"description": "...", "assignee": "..."\}.
  \end{minipage}
  \caption{Shared action-item-extraction prompt template. The outline prompt has the same scaffolding, with the instruction body replaced by a request for a markdown-heading outline of the meeting topics.}
  \label{fig:prompt-actions}
\end{figure*}

\paragraph{Enrollment-based speaker identification (L3).} Both stacks identify the speaker behind each diarization cluster by comparing a per-cluster representative audio clip against the workspace's enrollment clips and returning the matching display name. The OpenAI stack uses \texttt{gpt-audio} and the Gemini stack uses \texttt{gemini-3.1-pro-preview}; the system instruction and user-content scaffolding (Figure~\ref{fig:prompt-sid}) are essentially identical across the two stacks. The user content is multimodal: a short framing string, then one (\texttt{Enrollment --- \{display\_name\}:} + audio) pair per enrolled persona, then a final (\texttt{Target turn:} + audio) pair and a closed-list constraint. The response is a single line that we match back to the enrollment list with a tolerant first-name fallback.

\begin{figure*}[!t]
  \centering
  \begin{minipage}{0.97\textwidth}\small\ttfamily
\textbf{[System instruction]}\\
You identify which speaker uttered a target audio clip by comparing it to a small set of labeled enrollment clips. Respond with ONLY the label of the best-matching enrollment, exactly as given. Do not add punctuation, explanation, or commentary.\\[4pt]
\textbf{[User content, multimodal --- text + audio interleaved]}\\
Below are enrollment clips, each labeled with a speaker name, followed by a target turn clip. Identify which enrolled speaker produced the target clip. Reply with only the name.\\
Enrollment --- \{display\_name\_1\}: \textit{[audio Part]}\\
Enrollment --- \{display\_name\_2\}: \textit{[audio Part]}\\
\ldots\\
Target turn: \textit{[audio Part]}\\
Which enrollment label matches the target turn? Choose exactly one of: \{comma-separated display names\}.
  \end{minipage}
  \caption{Shared enrollment-based speaker-identification prompt (system instruction + user content scaffolding) used by both the OpenAI \texttt{gpt-audio} stage and the Gemini \texttt{gemini-3.1-pro-preview} stage at L3. The user content is multimodal: each \texttt{[audio Part]} placeholder is an inline audio attachment (the enrollment clip or the target turn).}
  \label{fig:prompt-sid}
\end{figure*}

\paragraph{Vocabulary post-correction (L3, OpenAI).} The OpenAI L3 stack runs a per-turn vocabulary-correction LLM pass over the upstream ASR transcript, prompted with the workspace's jargon list (acronyms, product names, person names); see Figure~\ref{fig:prompt-vocab-openai}. The response must be a JSON array of corrected text strings, length-equal to the input turns list; length-mismatched responses are rejected and the original turns are kept. The open-source SpeechBrain L3 stack (Appendix~\ref{sec:appendix:speechbrain}) reuses this exact prompt with \texttt{claude-haiku-4-5} as the rewriter, which is what drives the silent-summarization failure mode documented there.

\begin{figure*}[!t]
  \centering
  \begin{minipage}{0.97\textwidth}\small\ttfamily
You are correcting a meeting transcript for vocabulary errors only.\\[2pt]
EXPECTED VOCABULARY (terms likely spoken in this meeting):\\
\{vocab\}\\[2pt]
INSTRUCTIONS:\\
- Replace likely misspellings with the matching expected term (e.g., "B.L.E." \(\to\) "BLE", "Arc bus" \(\to\) "ArcBus", "Yuki Tanaka" stays as-is).\\
- Do NOT paraphrase, reorder, summarize, or fix grammar.\\
- Do NOT change speaker labels, turn boundaries, or punctuation beyond spelling fixes.\\
- Output ONLY a JSON array of strings, one entry per input turn, in the same order. Each entry is the corrected ``text'' for that turn.\\[2pt]
INPUT TURNS:\\
\{turns\_json\}
  \end{minipage}
  \caption{OpenAI L3 vocabulary post-correction prompt (\texttt{gpt-5.5}). The SpeechBrain L3 pipeline reuses this prompt verbatim with \texttt{claude-haiku-4-5} as the rewriter; on high-WER upstream input that pass tends to silently summarize around the transcript rather than patch it, which is the failure mode documented in Appendix~\ref{sec:appendix:speechbrain}.}
  \label{fig:prompt-vocab-openai}
\end{figure*}

\paragraph{Vocabulary post-correction (L3, Gemini).} The Gemini stack uses a stricter variant that adds explicit anti-insertion and anti-expansion guards on top of the OpenAI prompt (Figure~\ref{fig:prompt-vocab-gemini}). These constraints were added after we observed that the Gemini-flash rewriter is more aggressive than the OpenAI rewriter at applying vocabulary corrections to already-correct spans; the stricter prompt restores the intended ``vocabulary patch, not paraphrase'' behavior.

\begin{figure*}[!t]
  \centering
  \begin{minipage}{0.97\textwidth}\small\ttfamily
\textbf{[OpenAI prompt body, Figure~\ref{fig:prompt-vocab-openai}, then:]}\\[2pt]
ADDITIONAL CONSTRAINTS:\\
- The vocabulary list is REFERENCE ONLY. Most input tokens are already correct --- leave them alone.\\
- DO NOT replace a token just because a similar vocabulary entry exists. If the input token is already a valid English word, leave it unchanged.\\
- DO NOT change capitalization.\\
- DO NOT expand abbreviations or acronyms.\\
- DO NOT insert words. Output word count for each turn MUST be equal to or less than the input word count.\\
- If you are unsure whether a token is a misspelling, leave the turn unchanged. Conservative is correct.
  \end{minipage}
  \caption{Gemini-stack L3 vocabulary post-correction prompt. The instructions are the OpenAI body (Figure~\ref{fig:prompt-vocab-openai}) plus the explicit anti-insertion / anti-expansion constraints listed above.}
  \label{fig:prompt-vocab-gemini}
\end{figure*}

\paragraph{Adversarial-realism discriminator.} The structural-realism judge (\texttt{gemini-3.1-pro-preview}, Appendix~\ref{sec:appendix:refinement}) is shown a pairwise A/B comparison between the candidate and a length-matched AMI excerpt; its prompt (Figure~\ref{fig:prompt-discriminator}) explicitly scopes the verdict to discourse-level signals and bans the surface tells that the spoken-form formatter owns.

\begin{figure*}[!t]
  \centering
  \begin{minipage}{0.97\textwidth}\small\ttfamily
You are a forensic judge. You will see two meeting transcripts, labeled Transcript A and Transcript B. One is real; one is synthetic. Decide which is which based on structural signals only, then explain the tells that drove your decision.\\[2pt]
SCOPE --- structural signals ONLY. EXPLICITLY IGNORE: word-level disfluencies, single-word stutters, mid-word truncations, ASR transcription errors, diarization quirks ("Unknown Speaker", generic labels). A downstream pass owns those.\\[2pt]
CONSIDER (in-scope structural tells): (1) NARRATIVE ARC; (2) DROPPED THREADS; (3) ACTION-ITEM RECAP; (4) INTERRUPTION COST; (5) BANNED TROPES (mute, pets, deliveries, scripted forgetfulness, closing recap); (6) EXPOSITORY DIALOGUE; (7) SPEAKER ROLES; (8) ENDING.\\[2pt]
Transcript A:\\
\{\{TRANSCRIPT\_A\}\}\\[2pt]
---\\
Transcript B:\\
\{\{TRANSCRIPT\_B\}\}\\[2pt]
Reason briefly ($\le$500 chars) about the structural tells. Then output exactly one JSON object on the final line:\\
\{"verdict": "A"$\,|\,$"B", "reasoning": "\ldots", "structural\_issues": ["\ldots", \ldots]\}\\[2pt]
`verdict` is which transcript you believe is synthetic. `structural\_issues` is drawn from: clean\_arc $|$ no\_dropped\_threads $|$ explicit\_action\_items $|$ costless\_interruptions $|$ banned\_trope\_present $|$ expository\_dialogue $|$ clean\_speaker\_roles $|$ clean\_ending.
  \end{minipage}
  \caption{Adversarial-realism discriminator prompt (\texttt{gemini-3.1-pro-preview}). One transcript is the candidate, the other is a length-matched AMI excerpt; the synthetic side is randomized per (meeting, iteration) by a meeting-deterministic RNG so verdicts are reproducible.}
  \label{fig:prompt-discriminator}
\end{figure*}

\paragraph{Adversarial-realism refiner.} The refiner (\texttt{claude-opus-4-7}) is then told what the judge cited and asked to address those tells specifically. The full prompt (Figure~\ref{fig:prompt-refiner}) is long; we reproduce its key structural elements verbatim and elide ancillary boilerplate. The split between a same-family rewriter / refiner (Claude) and a cross-family discriminator (Gemini) is the empirical choice motivated in Appendix~\ref{sec:appendix:refinement}: same-family judges tend to declare the candidate indistinguishable on the first iteration even when obvious structural tells remain.

\begin{figure*}[!t]
  \centering
  \begin{minipage}{0.97\textwidth}\small\ttfamily
You are a transcript style editor. You previously rewrote a synthetic meeting transcript to look structurally real. A forensic judge correctly identified your rewrite as the synthetic one. Revise the transcript to directly address the tells the judge cited.\\[2pt]
\{\{LENGTH\_BUDGET\_CLAUSE\}\}\\[2pt]
SCOPE --- structural realism only (arc, dropped threads, unresolved blockers, action-item recap, interruption cost, banned tropes, expository dialogue, blurred speaker roles, trailing endings, structural disfluencies).\\[2pt]
OUT OF SCOPE --- do NOT add filler sounds, single-word stutters, or mid-word truncations (a downstream pass handles them).\\[2pt]
RULES: all key info preserved; do NOT add or remove speakers; you MAY split / merge / reorder turns within a small local window, or drop a short turn if "the moment didn't land".\\[2pt]
JUDGE'S REASONING: \{\{JUDGE\_REASONING\}\} \quad Structural issues flagged: \{\{STRUCTURAL\_ISSUES\}\}\\[2pt]
CHECKLIST (verify each before outputting): derail the main topic for 4+ turns at least once; add at least one place where someone talks past another; at least one speaker abandons a clause midway; when someone drops off, make it cost something; remove every banned interruption trope; replace expository dialogue with shorthand; blur at least one speaker role; trail off the ending with logistics --- NO end-of-meeting recap.\\[2pt]
Real transcript style reference: \{\{REAL\_REFERENCE\}\}\\
Previous rewrite: \{\{PREVIOUS\_REWRITE\}\}\\[2pt]
Self-check (mandatory): identify the single most synthetic-looking feature of your current draft and fix it before outputting. Return ONLY a JSON array of turns of shape \{"speaker\_id": "\ldots", "speaker\_name": "\ldots", "text": "\ldots"\}.
  \end{minipage}
  \caption{Adversarial-realism refiner prompt (\texttt{claude-opus-4-7}). The discriminator's per-iteration verdict reasoning and structural-issue tags are spliced into the prompt so the refiner addresses the cited tells directly; the CHECKLIST encodes the structural realism targets used throughout the corpus.}
  \label{fig:prompt-refiner}
\end{figure*}

\paragraph{Summary-quality judge.} The \textsc{faithfulness} and \textsc{coherence} scores reported in Tables~\ref{tab:headline} and~\ref{tab:isomorphic-vs-standard} are produced by the prompt in Figure~\ref{fig:prompt-summary-judge}, executed by \texttt{claude-sonnet-4-6} with content-hash caching (\S\ref{sec:experiments:metrics}). The prompt elicits five dimensions in a single pass; we report these two and release the remainder with the corpus. The prompt is a direct-scoring LLM-as-judge in the spirit of \citet{zheng-etal-2023-judging}: the judge writes integer 1--5 scores directly into a JSON object that we parse, with hand-written per-dimension rubric anchors (we do not use auto-generated chain-of-thought steps or token-probability-weighted scoring as in \citealp{liu-etal-2023-g-eval}). The judge is reference-grounded --- inputs are the gold transcript, the gold summary, the workspace roster, and the system summary --- rather than scoring the system summary in isolation. Three design choices in the prompt are load-bearing for how Tables~\ref{tab:headline} and~\ref{tab:isomorphic-vs-standard} should be read. (i) The prompt explicitly decouples length from every dimension except conciseness, so longer or shorter summaries are not rewarded under recall, precision, faithfulness, or coherence; reference length is not a target. (ii) The prompt marks any workspace-canonical-name expansion that is consistent with the roster (e.g.\ \emph{Brindle} $\to$ \emph{Brindlewood Commerce Group}, first name $\to$ full name, codename $\to$ full project description) as faithful rather than as a fabrication; the alternative would penalize precisely the L3 behavior the experiment is meant to measure, and is what drives the Gemini L3 faithfulness dip on stacks whose roster-expansion behavior is more aggressive (\S\ref{sec:results:metrics}). (iii) Speaker attribution is excluded from faithfulness, since a separate metric (\textsc{cross-meeting speaker-identity consistency}) scores it; speaker labels are treated as opaque tags by the judge. We reproduce the prompt body verbatim and abridge only the per-dimension Likert-anchor rubrics.

\begin{figure*}[!t]
  \centering
  \begin{minipage}{0.97\textwidth}\small\ttfamily
You are evaluating the quality of a meeting summary generated by a speech-to-text system.\\[2pt]
You are given: (1) the WORKSPACE CONTEXT (canonical roster of company, people, and projects --- the source of truth for full canonical names, affiliations, and project descriptions); (2) the REFERENCE TRANSCRIPT; (3) the REFERENCE SUMMARY; (4) the SYSTEM SUMMARY.\\[2pt]
WORKSPACE CONTEXT: \{workspace\_context\} \quad REFERENCE TRANSCRIPT: \{transcript\} \quad REFERENCE SUMMARY: \{ref\_summary\} \quad SYSTEM SUMMARY: \{sys\_summary\}\\[2pt]
Evaluate the SYSTEM SUMMARY on five dimensions, scoring each from 1 (worst) to 5 (best). Use the full 1--5 range --- pick 2 or 4 when the summary sits between the anchors below.\\[2pt]
IMPORTANT --- length is not quality. The system summary may be substantially shorter or longer than the reference; reference length is not a target. Judge each dimension on what the content does, not how much text it occupies:\\
- Recall: count distinct important facts covered, not bullets, paragraphs, or word count. Extra prose that does not add new facts does not raise recall.\\
- Precision: a short summary that includes only important points scores 5; do NOT penalize it for being brief (omissions are recall's job, not precision's).\\
- Faithfulness: score only on whether claims are supported by the transcript and/or workspace context. Do not reward length, polish, or apparent ``thoroughness.''\\
- Coherence: score only on structure and flow. Do not reward longer summaries for appearing more comprehensive.\\
- Conciseness is the ONLY dimension where length directly matters.\\[2pt]
\textbf{Faithfulness (1--5):} Evaluate factual content only: events, decisions, action items, deadlines, numbers, technical claims, products / projects / features, dependencies, and outcomes. IGNORE speaker attribution. Treat speaker labels (real names, ``Speaker 1'', etc.) as opaque tags --- do NOT check whether a statement is assigned to the right person. Speaker-attribution quality is measured by a separate metric and is out of scope here.\\
Workspace-consistent entity resolution is faithful, NOT a fabrication. When the summary expands a short form to its canonical workspace name (e.g., \emph{Brindle} $\to$ \emph{Brindlewood Commerce Group}, first name $\to$ full name, abbreviation $\to$ full project / product name, project codename $\to$ full project description) and that expansion matches the WORKSPACE CONTEXT, treat it as accurate. Only flag entity names or facts that are absent from BOTH the transcript and the workspace context, or that conflict with the workspace context.\\[2pt]
\emph{[Per-dimension 1--5 anchor rubrics for Recall, Precision, Faithfulness, Coherence, Conciseness follow the standard G-Eval pattern \citep{liu-etal-2023-g-eval}; elided here for space.]}\\[2pt]
Respond with ONLY a JSON object: \{\{\,``recall'':\,\textless int\textgreater, ``precision'':\,\textless int\textgreater, ``faithfulness'':\,\textless int\textgreater, ``coherence'':\,\textless int\textgreater, ``conciseness'':\,\textless int\textgreater, ``recall\_explanation'':\,\textless brief\textgreater, ``precision\_explanation'':\,\textless brief\textgreater, ``faithfulness\_explanation'':\,\textless brief\textgreater, ``coherence\_explanation'':\,\textless brief\textgreater, ``conciseness\_explanation'':\,\textless brief\textgreater\,\}\}
  \end{minipage}
  \caption{Summary-quality LLM-judge prompt (\texttt{claude-sonnet-4-6}). Reference-grounded direct-scoring LLM-as-judge in the spirit of \citet{zheng-etal-2023-judging}, with hand-written per-dimension 1--5 rubric anchors; we do not use the auto-generated chain-of-thought steps or token-probability-weighted scoring of \citet{liu-etal-2023-g-eval}. The judge is explicitly instructed to (i) treat roster-consistent canonical-name expansions as faithful rather than as fabrications, (ii) decouple length from every dimension except conciseness, and (iii) exclude speaker attribution from faithfulness (scored separately). Per-dimension Likert anchors are abridged for space; full prompt is released with the evaluation code.}
  \label{fig:prompt-summary-judge}
\end{figure*}

\section{Qualitative error examples}
\label{sec:appendix:error-examples}

To make the error categories in \S\ref{sec:errors} concrete, we walk through representative spans drawn from the OpenAI system on \textsc{OSynth} at L1, L2, and L3, all taken from a single pilot meeting.

\paragraph{Transcript-level TERM recovery.} The same passage shows progressively cleaner handling of hardware-domain terms as workspace context is added (see Table~\ref{tab:qual-terms} for the contrast):

\begin{quote}\small
\textbf{GT:} \emph{Yuki:} Already doing it. PMIC and BLE, nothing there needs the seven-one-four-two. [\ldots] The pipeline's tied to the timing and resolution profile---ArcBus gives us some abstraction, but the digital filter coefficients are dictated by the analog characteristics of the front-end.

\textbf{L1:} \emph{Unknown speaker:} Already doing it, PMIC and VLE Nothing there needs the seven one four two. [\ldots] ARC plus gives us um some abstraction\ldots

\textbf{L2:} \emph{Unknown speaker:} Already doing it. PMIC and PLE Nothing there Needs the SN7142. [\ldots] Arctus gives us um some abstraction\ldots

\textbf{L3:} \emph{Darren Okafor:} Already doing it, PMIC and BLE Nothing there needs the seven one four two. [\ldots] ArcBus gives us um some abstraction\ldots
\end{quote}

\begin{table}[!t]
  \centering
  \small
  \setlength{\tabcolsep}{4pt}
  \begin{tabular}{@{}lllll@{}}
    \toprule
    \textbf{Term} & \textbf{GT} & \textbf{L1} & \textbf{L2} & \textbf{L3} \\
    \midrule
    BLE    & BLE    & VLE      & PLE    & BLE \checkmark \\
    ArcBus & ArcBus & ARC plus & Arctus & ArcBus \checkmark \\
    \bottomrule
  \end{tabular}
  \caption{Progressive resolution of TERM errors (\textsc{BLE}, \textsc{ArcBus}) across L1, L2, and L3 from a single passage. L2's self-biased prior outputs recover some tokens (\textsc{PMIC}, \textsc{SN7142}) but also introduce confident hallucinations (\textsc{Arctus}); the explicit L3 jargon list recovers the remainder.}
  \label{tab:qual-terms}
\end{table}

\paragraph{Summary-level NE recovery.} The same NE / TERM failures propagate into the summary. Excerpts from the same meeting's L1 vs.\ L3 summaries:

\begin{quote}\small
\emph{L1 (excerpt):} \textbf{Sony Tech SN714 slip / EVT impact.} Two SN714 samples have slipped by at least three weeks due to fab allocation. Team is treating three weeks as a minimum delay\ldots

\emph{L3 (excerpt):} \textbf{Sonitec SN7142 supply slip / EVT impact.} Darren flagged that Sonitec SN7142 samples have slipped at least three weeks; team should treat that as a floor, not an optimistic estimate. Switching to a second source was judged likely worse than waiting: AFE redesign, new HAL driver work, and re-characterization. Outcome: continue work not dependent on the SN7142---Gen 2 bring-up, PMIC, BLE, firmware.
\end{quote}

L3 correctly names \emph{Sonitec SN7142} (vs.\ L1's \emph{Sony Tech SN714}), attributes the flag to \emph{Darren} by name, and adds the explicit outcome decision absent from L1.

\paragraph{Action-item assignee correction.} The transcript-level NE failures also drive incorrect attribution in extracted action items. From the same meeting:

\begin{quote}\small
\textbf{GT:} \emph{Darren:} Deliver Arc enclosure stack-up by Friday. \\
\textbf{L1:} \emph{[Ryan]} Deliver the odd enclosure power stack-up by Friday. \\
\textbf{L3:} \emph{[Darren Okafor]} Deliver the Arc enclosure tolerance stack-up by Friday.
\end{quote}

L1 hallucinates \emph{Ryan} (a name not present in the meeting) and corrupts the artifact name (\emph{odd} for \emph{Arc}); L3's enrollment-based SID and jargon list jointly recover both the assignee and the artifact.

\paragraph{NUM recovery (number recall, +0.20).} The +0.20 absolute jump in number recall (\S\ref{sec:results:headline}) is driven by L3's ability to anchor ASR-fragile identifiers against the workspace text artifacts. A representative span from the \emph{Marcus \(\leftrightarrow\) Simone} 1:1 (2026-05-19, fintech-payments workspace) discusses ISO~20022 message-family identifiers (\texttt{pacs.008} and \texttt{camt.054}), which are spelled out by the active project spec under the heading ``ISO 20022 Payment Message Migration'':

\begin{quote}\small
\textbf{GT:} \emph{PACS.008 is decent. Three of the CAMT.054 sub-fields are still draft, Payments Canada hasn't published.}

\textbf{L1:} \emph{PAX eight is decent, three of the CAMTWOH five four sub-fields are uh still draft Payments Canada hasn't hasn't published.}

\textbf{L3:} \emph{pacs.008 is decent. Three of the camt.054 subfields are, uh, still draft Payments Canada hasn't hasn't published.}
\end{quote}

L1 parses \texttt{pacs.008} as ``\textit{PAX eight}'' (a phonetic guess that drops both the prefix and the dotted subtype) and \texttt{camt.054} as ``\textit{CAMTWOH five four}'' (a syllable-level mangling of the spelled-out form). L3 recovers both via the workspace jargon list and project-spec excerpt, which contain the canonical surface forms. The per-meeting number-recall score moves from $0.00$ at L1 to $1.00$ at L3 on this meeting; the L2 hypothesis already recovers both identifiers via the vocab post-correction LLM pass even without the workspace name list, consistent with the ablation finding (\S\ref{sec:results:ablations}, Appendix~\ref{sec:appendix:ablations}) that workspace text---not enrollments---drives the number-recall axis.

\paragraph{Assignee correction, second meeting.} To rule out a single-meeting cherry-pick of the assignee gain, we give a second OpenAI assignee example from a different meeting in the same workspace---the \emph{Marcus \(\leftrightarrow\) Devon} 1:1 (2026-05-19)---where the canonical assignee is \emph{Devon Ashby} (workspace person \texttt{p8}). The L1, L2, and L3 hypotheses trace out an \emph{invented name $\to$ invented name $\to$ canonical} trajectory rather than a flat accuracy lift:

\begin{quote}\small
\textbf{GT:} \emph{[Devon Ashby]} Document idempotency-key behavior during failover/retry overlap, including whether the key survives a gateway switch, in the CAS ADR. \\
\textbf{L1:} \emph{[Bevan]} Trace the idempotency key behavior during failover, especially when a CAS retry overlaps a gateway switch, to confirm whether duplicate charging is possible. \\
\textbf{L2:} \emph{[Devin]} Trace the idempotency-key behavior when a retry overlaps a gateway failover transition, including what the downstream sees after the gateway switch. \\
\textbf{L3:} \emph{[Devon Ashby]} Trace idempotency-key behavior during failover/retry overlap, including whether the key survives a gateway switch and what the downstream sees.
\end{quote}

The L1 SID stage has no workspace name list, so it autocompletes the diarization cluster to a phonetically plausible but invented English name (\emph{Bevan}); the L2 vocab-correction pass nudges the guess toward a different invented neighborhood (\emph{Devin}) but still cannot resolve to any canonical workspace identity. Only at L3 do the workspace name list and the enrollment-based SID jointly anchor the cluster to person \texttt{p8} (\emph{Devon Ashby}). The per-meeting assignee accuracy is $0.00$ at L1 and $1.00$ at L3, and the same pattern holds across six near-identical Devon-assigned action items in the meeting, so the correction is not a single-item cherry-pick.

\paragraph{Cross-meeting speaker consistency (+0.26).} The cross-meeting speaker-identity consistency metric (\S\ref{sec:experiments:metrics}) flips most visibly when the same person appears in two near-consecutive meetings whose per-meeting diarization clusters carry incompatible labels. Picking up the same \emph{Devon Ashby} (\texttt{p8}) as in the assignee example above, Table~\ref{tab:qual-cross-meeting} shows the labels assigned by the OpenAI L1 versus L3 systems in two meetings two days apart in the fintech-payments workspace:

\begin{table}[!t]
  \centering
  \footnotesize
  \setlength{\tabcolsep}{5pt}
  \renewcommand{\arraystretch}{1.1}
  \begin{tabular}{@{}l l l@{}}
    \toprule
                          & \textbf{Meeting $n$} & \textbf{Meeting $n{+}1$} \\
    \midrule
    Title                 & Marcus--Devon 1:1    & BDW Sandbox Scope \\
    Date                  & 2026-05-19           & 2026-05-21 \\
    \addlinespace[2pt]
    L1 label              & \emph{Bevan}         & \emph{Devin} \\
    L3 label              & \emph{Devon Ashby}   & \emph{Devon Ashby} \\
    \bottomrule
  \end{tabular}
  \caption{Cross-meeting labels for canonical person \texttt{p8} (\emph{Devon Ashby}) in two near-consecutive meetings in the same workspace, under OpenAI L1 versus L3. L1 hallucinates a different phonetically plausible English given name in each meeting because per-meeting diarization clusters do not persist and the L1 SID stage has no workspace name list to anchor on; L3 fixes both meetings to the same canonical identity via the workspace name list and enrollment-based SID against the persona's prompt clip.}
  \label{tab:qual-cross-meeting}
\end{table}

The two L1 labels disagree both with the canonical name and with each other---a single-meeting accuracy metric would mark them merely as ``wrong'', but the cross-meeting consistency metric exposes the structurally worse failure that no downstream consumer of the L1 output can stitch the two meetings into a single \emph{Devon} entity. L3 recovers consistency on both axes because the per-meeting cluster-to-canonical-identity step is anchored to the same workspace name list and the same enrollment clip across meetings (Appendix~\ref{sec:appendix:workspace-walkthrough}).

\section{TTS rendering strategy}
\label{sec:appendix:tts-rendering}

Recent multi-speaker TTS systems can render an entire multi-party conversation in a single forward pass, taking as input a single script with interleaved speaker tags and emitting one mixed waveform together with implicit turn timings. We deliberately do \emph{not} use this single-call rendering mode in the released pipeline. The convenience of a one-shot call is offset by three concrete losses that matter for downstream evaluation: (i) the absence of an explicit per-turn audio segment makes word- and turn-level alignment back to the canonical transcript ambiguous, and any subsequent speaker-attributed reference (RTTM, word-level alignments, per-speaker enrollments) has to be re-derived by a forced-alignment pass with its own error budget; (ii) per-turn acoustic parameters that vary by persona (rate, pitch range, vocal effort, accent) are difficult to control independently when all turns are produced in a single decoding pass, which compresses the cross-speaker acoustic diversity that diarization and speaker-identification systems are meant to be tested on; and (iii) when a single decoding pass mishandles one turn (a degenerate repetition, a swallowed word, an off-target voice), the whole meeting must be re-rendered, whereas a per-turn pipeline localizes the failure to a single turn that can be regenerated independently. We therefore synthesize each turn independently (\S\ref{sec:dataset:rendering}) with the persona's reference clip, the persona's target acoustic parameters, and the turn's TTS-form text, and assemble the mixed waveform from the resulting per-turn segments. Per-turn segments are retained alongside the mixed waveform so that downstream users can re-mix, re-spatialize, or compose individual turns into new scenarios.

The per-turn pipeline does introduce its own failure mode: voice-cloning prompts drawn from a third-party corpus occasionally trigger degenerate behavior in Qwen3-TTS \citep{qwen3-tts-2025}, including stuck loops, silence, or off-target speaker identity. We mitigate this by pre-processing the CommonVoice \citep{ardila-etal-2020-common-voice} reference pool offline before any meeting is rendered: each candidate clip is passed through an ASR round-trip check against its own transcript, a duration and SNR filter, and a short stability probe in which the clip is used to synthesize a fixed neutral utterance which is then re-transcribed; clips for which the round-trip WER exceeds a threshold, or for which the stability probe degenerates, are dropped from the pool. The surviving pool, which we refer to internally as the voice cache, is the only set of clips that personas are sampled from. This preprocessing turns a per-meeting failure mode into a one-time pool-curation cost and, in practice, eliminates the long-tail TTS failures that we observed when sampling reference clips on the fly.

For inference at corpus scale, we serve the 1.7B-parameter variant of Qwen3-TTS \citep{qwen3-tts-2025} on a 32$\times$A100~80\,GB cluster using batched per-turn synthesis through a \texttt{vllm-omni} backend, ultimately sustaining a real-time factor of approximately 0.5 (one second of synthesized speech per half-second of wall-clock time across the batched job). This throughput is what makes rendering the 570 hours of audio in the released corpus tractable in a single pass and what makes the regeneration cost of individual failed turns (rather than whole meetings) negligible in practice.

\section{Persona-to-voice assignment}
\label{sec:appendix:voice-assignment}

Voices are assigned from the curated pool of Appendix~\ref{sec:appendix:tts-rendering} by matching a single attribute: gender. Each persona carries a gender assigned by the LLM during workspace sampling (\S\ref{sec:dataset:pipeline}), and the curated Common Voice pool carries a gender label per voice ID (\emph{male}, \emph{female}, or unknown). Personas are partitioned by their assigned gender and drawn from the corresponding subset of voice IDs; a persona whose gender is non-binary or unset draws from the union of all subsets. Within the matching subset the draw is uniform and without replacement inside a workspace, so no two personas in a workspace share a voice. No other persona attribute (age, seniority, role, accent, or communication style) currently conditions the assignment, which is a deliberate simplification rather than a modelling claim: richer demographic alignment would require voice-pool metadata that Common Voice does not consistently provide.

\section{Conversational prosody}
\label{sec:appendix:prosody}

Two prosodic phenomena that are central to natural conversational audio (turn-taking gaps and backchannels) are produced by the pipeline rather than left to the TTS system. We give them an explicit treatment because they are easy to omit and because both of them strongly affect how realistic the mixed waveform sounds to a human listener and how diarization systems behave on it.

\paragraph{Floor-transfer offsets.} For each consecutive speaker change we sample a floor-transfer offset (FTO), the signed gap between the end of speaker $A$'s turn and the start of speaker $B$'s turn, from a mixture distribution fit to publicly available conversational corpora; complementary lines of work model turn-taking sequences themselves as a function of team-member trait composition \citep{obryan-etal-2024-mlspeak}, whereas we leave who-speaks-next to the LLM-driven turn-generation pass (\S\ref{sec:dataset:pipeline}) and target only the timing of the resulting transitions here. The mixture concentrates short positive gaps in the range typically reported for spontaneous multi-party meetings, with a tail of longer gaps to model topic-boundary pauses and a negative-gap component that produces brief overlaps at turn boundaries. Same-speaker continuations (a single speaker holding the floor across multiple turns) draw from a separate distribution dominated by short intra-turn pauses. The exact distributional parameters used for the released corpus are recorded alongside the audio and can be re-fit if a user wants to target a different conversational register.

\paragraph{Substantive turn-level overlap.} Beyond the small negative-gap overlaps produced at turn boundaries by the FTO distribution, an LLM annotator that runs after the spoken-form pass scans the formatted turn list and emits, for each non-first turn, an optional \emph{overlap anchor} pointing at a specific word of the immediately preceding turn together with a mode label: \textsc{interrupt}, where the anchor turn is truncated at the anchor word and the overlapper takes the floor; or \textsc{concurrent}, where the anchor turn continues to completion while the overlapper speaks over it. Audio placement is two-pass: non-overlap turns are placed left-to-right by sampled FTO from the previous non-overlap turn's offset, then overlap turns are placed at the anchor word's forced-alignment onset and the cumulative timeline pointer is not advanced. A final per-speaker monotonic sweep iterates to a fixed point, dropping any annotation that would cause a speaker to overlap with themselves, so the released audio satisfies a per-speaker no-self-overlap invariant by construction (further checked in the validation stage, Appendix~\ref{sec:appendix:validation}). Annotator failures (malformed payloads, exceptions) fall back to FTO placement; the audio pipeline never blocks on overlap.

\paragraph{Backchannels.} Single-token backchannels (\emph{mm-hm}, \emph{yeah}, \emph{right}) are generated as ordinary turns by the transcript LLM and then identified and placed by two dedicated annotators that run after the spoken-form pass. A sparse classifier scans the formatted turn list and emits, for each turn it judges to be a backchannel, an explicit \texttt{bc\_kind} label drawn from \{\textsc{agreement}, \textsc{acknowledgment}, \textsc{continuer}, \textsc{reaction}\}; substantive turns are non-BC by absence. A second pass anchors each identified backchannel to the most recent prior non-backchannel, non-overlap turn and forces \texttt{mode=concurrent} so the backchannel overlays the floor-holder's audio without advancing the meeting clock. Per-speaker backchannel priors (each persona's vocabulary of customary backchannel phrases) condition the classifier so that backchannel forms vary across speakers. Backchannel rows are preserved in the transcript and RTTM under the original speaker's identity, with an \texttt{is\_backchannel} flag that allows downstream metrics to filter them in or out (e.g., opt-in WER scoring). The classifier and anchor pass each have heuristic fallbacks on truncation or parse failures and never block the pipeline.

\section{Speech pacing relative to the cloning prompts}
\label{sec:appendix:pacing}

The reference clips used for voice cloning are spontaneous speech (Appendix~\ref{sec:appendix:licenses}), yet the rendered turns can still sound read. To quantify how much of this comes from the TTS stage rather than the prompts, we ran the NeMo forced aligner \citep{rastorgueva-etal-2023-nfa} over both the cloning prompts and the generated turns and macro-averaged per speaker (Table~\ref{tab:pacing}). \textsc{OSynth} turns are spoken faster and, more importantly, are paced far more uniformly: the inter-word pause ratio falls by 60\% and pauses longer than 200\,ms become roughly three times rarer per minute. We attribute the residual read-aloud character primarily to this pacing uniformity, and secondarily to a domain mismatch, since even spontaneous Common Voice speech is recorded in isolation in response to prompts rather than in live multi-party conversation. We expect this effect to diminish as TTS models improve.

\begin{table}[!t]
  \centering
  \footnotesize
  \setlength{\tabcolsep}{4pt}
  \begin{tabular}{lrrr}
    \toprule
    \textbf{Measure} & \textbf{Prompts} & \textbf{\textsc{OSynth}} & \textbf{$\Delta$} \\
    \midrule
    Active-speech WPM        & 134.5 & 180.9 & $+34.5\%$ \\
    Pause ratio              & 47.9\% & 18.9\% & $-60.5\%$ \\
    Pauses $>$200\,ms / min  & 44.5 & 15.6 & $-65.0\%$ \\
    \bottomrule
  \end{tabular}
  \caption{Pacing of the voice-cloning prompts versus the rendered \textsc{OSynth} turns, from NeMo forced alignments macro-averaged per speaker. The generated turns are faster and markedly more uniform in pausing.}
  \label{tab:pacing}
\end{table}

\section{Alignment procedure and validation}
\label{sec:appendix:alignment}

Every released meeting ships with word-level and turn-level alignments back to the canonical transcript (\S\ref{sec:dataset:rendering}); these alignments are what make the RTTM file, the per-speaker enrollments, and the contextualized-transcription metrics of \S\ref{sec:experiments:metrics} well-defined. Because each turn is rendered independently from a known TTS-form string, we obtain initial per-word timings directly from the TTS aligner that ships with Qwen3-TTS \citep{qwen3-tts-2025}, anchored to the turn's offset in the mixed waveform. We then refine these per-turn timings with a forced-alignment pass against a small open-source acoustic model on the unmixed per-turn audio, and propagate the refined timings through the mixing offsets and the pyroomacoustics convolution (Appendix~\ref{sec:appendix:audio-postproc}); the convolution introduces a known per-speaker delay that is subtracted analytically rather than re-estimated.

To validate the alignment quality, we evaluate on a held-out sample of meetings against per-utterance reference timings derived from the per-turn rendering offsets (which are exact by construction). We report onset mean absolute error, onset RMSE, duration mean absolute error, and word-coverage rate. We additionally measure the rate at which a word's predicted span lies entirely within the correct speaker's RTTM interval, which is the property the speaker-attributed evaluation cares about. Across the validation sample, onset MAE is well below 0.25\,s and word-coverage exceeds 99\%; we report the exact numbers and the validation procedure in the released code repository.

\section{Audio post-processing}
\label{sec:appendix:audio-postproc}

After per-speaker TTS synthesis and mixing (\S\ref{sec:dataset:rendering}), each meeting is passed through a \texttt{pyroomacoustics} \citep{scheibler-etal-2018-pyroomacoustics} post-processing stage that places the speakers in a simulated environment and renders the recording as if it had been captured by one or more microphones. For every meeting we sample a shoebox room geometry (length, width, and height drawn from ranges typical of small-to-medium meeting rooms), a wall absorption coefficient drawn from a range consistent with carpeted, lightly furnished office spaces, and a microphone position near the center of the table. Each persona is assigned a fixed seat around the table for the duration of the meeting; we then compute the image-source-model impulse response from each seat to the microphone and convolve each speaker's TTS turns with the corresponding response before mixing the convolved turns onto the final track. The resulting waveform exhibits reverberation, inter-speaker level differences, and spectral coloration consistent with a shared microphone recording, while preserving the exact word-level alignment back to the canonical transcript. The geometry, seat layout, and absorption parameters used for each meeting are released alongside the audio.

\paragraph{Hybrid (in-room plus remote) calls.} A substantial fraction of workspace meetings are not fully co-located: one or more attendees join over a video-conferencing client from a separate room, a home office, or a phone. To cover this case, a sampled fraction of \textsc{OSynth} meetings are rendered in a \emph{hybrid} configuration in which a subset of attendees is placed inside the shared conference room as above, while the remainder are rendered as remote participants. Each remote participant is placed in their own (smaller, more reverberant or more dampened) room with its own microphone, convolved with that room's impulse response, and then passed through a lightweight codec / bandwidth simulation that approximates the spectral compression and occasional level mismatch of a typical VoIP downlink before being mixed into the shared track at a per-participant gain. In-room and remote participants therefore arrive at the final mix with markedly different reverberation profiles and bandwidths, which is the property that distinguishes hybrid recordings from fully co-located ones and that downstream diarization and speaker-identification systems must contend with in deployment. The hybrid-vs-co-located flag, the per-participant room assignment, and the codec configuration are released alongside each meeting so that users can subset the corpus by acoustic condition.

\section{Pipeline validation}
\label{sec:appendix:validation}

Every persisted artifact in the pipeline is validated against a typed schema both at generation time and again on read, which catches malformed LLM outputs (\texttt{null} action items, missing fields, nicknames emitted in identifier slots) at the boundary they originate from rather than corrupting downstream metrics silently. A final pipeline-validation stage additionally checks a small set of cross-artifact invariants on every released meeting: every active attendee has a join mode consistent with the meeting's mode (\emph{virtual} meetings reject in-room participants, \emph{in-person} meetings reject remote participants, and \emph{hybrid} meetings with two or more active attendees require at least one of each); every meeting's audio length is within a fill-ratio band of its scheduled calendar slot (\S\ref{sec:dataset:pipeline}), with under-filled meetings flagged in an aggregate \texttt{duration\_fill.json}; and no speaker overlaps with themselves in the released RTTM (Appendix~\ref{sec:appendix:prosody}), with violations counted in an aggregate \texttt{overlap\_stats} block. The validator emits warnings rather than aborting the run, but the aggregate counts and per-meeting flags are released alongside the corpus so that users can subset out any meeting that fails an invariant.

\section{Multilingual generation}
\label{sec:appendix:multilingual}

The \textsc{OSynth} pipeline (\S\ref{sec:dataset:pipeline}--\ref{sec:dataset:rendering}) is language-agnostic: the LLM prompts that produce the latent state and turn-by-turn dialogue accept a target language as an argument, and the TTS stage uses Qwen3-TTS \citep{qwen3-tts-2025}, which supports voice cloning across all of its training languages. To verify this, using language-matched voice pools and the same demographic-matching procedure as the English release, we generated workspaces in five additional languages: German, Spanish, French (reference voices from CommonVoice \citep{ardila-etal-2020-common-voice}), Portuguese (CORAA \citep{candido-junior-etal-2022-coraa}), and Korean (KsponSpeech \citep{bang-etal-2020-ksponspeech}). We do not include these workspaces in the main corpus or evaluation: the human-discrimination study (\S\ref{sec:dataset:realism}) is English-only, so we cannot make comparable realism claims for the non-English data. The main release is therefore English-only; we expect to release non-English workspaces in a follow-up once the corresponding validation has been run.

\section{Annotation interface}
\label{sec:appendix:annotation-ui}

\begin{figure*}[!t]
  \centering
  \includegraphics[width=0.95\textwidth]{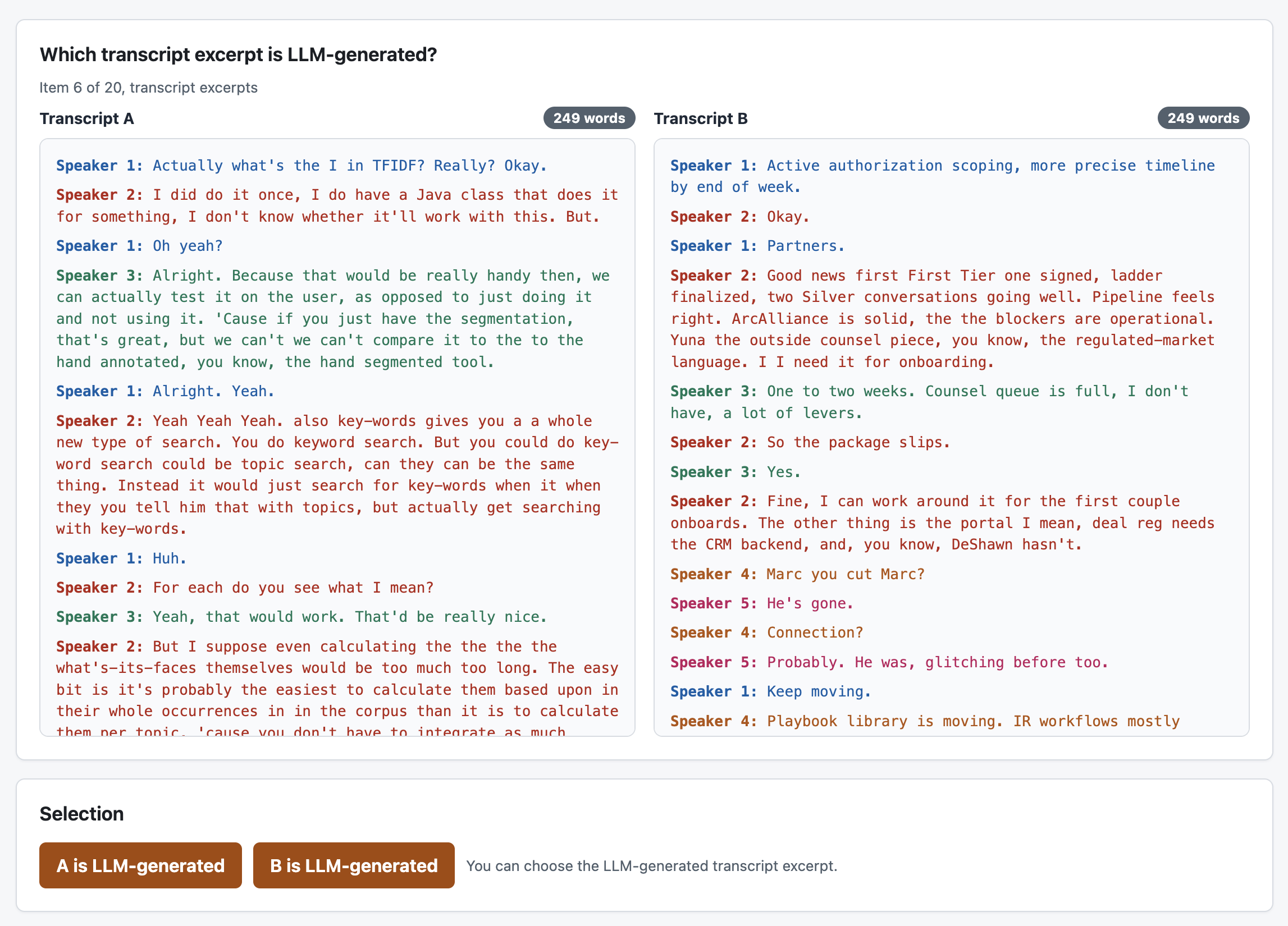}
  \caption{Web interface used in the 2AFC human-discrimination study (\S\ref{sec:dataset:realism}). Each trial displays a pair of length-matched transcript excerpts (one real, one synthetic), with speaker labels anonymized to generic identifiers and a word-count badge above each excerpt; the annotator selects which excerpt is LLM-generated using the two-button control at the bottom. The transcript textboxes are scrollable.}
  \label{fig:annotation-ui}
\end{figure*}

Figure~\ref{fig:annotation-ui} shows the recorded-trial view of the web interface used by annotators in the human-discrimination study (\S\ref{sec:dataset:realism}). Annotators were recruited from a pool of professionals working in the meeting-transcription domain. Before the recorded trials began, each session presented a written instructions panel followed by three practice trials with immediate correctness feedback (an extreme attention-check-style fake, a realistic \textsc{OSynth} fake, and a Haiku-w/o-obfuscation fake, each paired against the same real excerpt); practice responses were not stored. The verbatim instructions text is reproduced below. Each recorded trial then presents two length-matched transcript excerpts side by side (one real, one synthetic), with speaker labels replaced by generic identifiers (\emph{Speaker 1}, \emph{Speaker 2}, \ldots) to prevent name-based shortcuts and a word-count badge above each excerpt to make the length-matching transparent. A two-button forced-choice control beneath the excerpts records the annotator's selection of which excerpt is LLM-generated. Trial progress (e.g., ``Item 6 of 20'') is shown at the top of the panel, and attention-check trials are interleaved with realistic and Haiku-w/o-obfuscation trials in a fixed per-session order.

\paragraph{Instructions shown to annotators.} The instructions panel preceding the practice and recorded trials read, verbatim:

\begin{quote}\small
Each item shows two short transcript excerpts of similar length. One excerpt is actual meeting audio manually transcribed by a person. The other is an LLM-generated transcript or a deliberately modified attention-check transcript.

Your job is to select the LLM-generated transcript.

Do not decide from domain terms, project names, or acronyms alone. An actual meeting can involve a fictional or roleplayed task, and an LLM-generated transcript can include realistic technical vocabulary.

Some recorded items are attention checks. If one option clearly contains task-aware or deliberately edited text, select that option.

The examples below are practice only and are not saved. You will get immediate feedback before the recorded trials begin.

You should spend no more than 2 minutes on each recorded item. You can choose after 20 seconds.
\end{quote}

\section{Nicknames}
\label{sec:appendix:nicknames}

Real workspace conversation rarely uses canonical full names for familiar colleagues, and ASR systems whose post-correction is biased toward an org-chart roster of canonical names will systematically miss the informal forms that actually appear on the audio. To make this dimension measurable, every persona in \textsc{OSynth} carries an explicit nickname policy that is propagated through both the latent state and the rendered transcript.

Each persona record carries two nickname fields in addition to its canonical \texttt{name}: an optional \texttt{nickname} string and a \texttt{nickname\_usage} policy drawn from \{\texttt{none}, \texttt{sometimes}, \texttt{mostly}\}. The distribution is controlled at workspace-sampling time by the people-sketch prompt: most personas receive \texttt{none}, a minority receive a plausible informal nickname with \texttt{sometimes}, and a smaller minority receive one with \texttt{mostly}. The transcript-generation prompt receives the policy as part of the attendee roster and is asked to vary how other speakers refer to a person accordingly: \texttt{mostly} personas are referred to by nickname in nearly all informal turns, \texttt{sometimes} personas appear under a natural mix of canonical and informal forms, and \texttt{none} personas always appear by canonical name.

Identity remains anchored to the stable \texttt{person\_id} throughout the pipeline: the transcript LLM only emits \texttt{(speaker\_id, text)} pairs, and the human-facing \texttt{speaker\_name} label is derived downstream (nickname when \texttt{usage = mostly}, canonical otherwise). Downstream artifacts (summaries, action items, emails, documents) continue to be addressed by the same \texttt{person\_id}, so a nickname is never used as an alias key in the data model.

This separation is what keeps name scoring well-defined under either convention. The release carries both the canonical names from the org chart and persona profile and whichever forms (canonical or nickname) actually appear in the oracle transcript for a given meeting, so a scorer can be required to match either. The two diverge most for personas with \texttt{usage = mostly}, where a system that recovers only stored names misses the forms that were actually spoken.

\section{Gemini: isomorphic vs.\ standard metrics}
\label{sec:appendix:gemini-iso-std}

Table~\ref{tab:isomorphic-vs-standard-gemini} reports the L1$\to$L3 breakdown of isomorphic and standard metrics for the Gemini stack, complementing the OpenAI-side breakdown in Table~\ref{tab:isomorphic-vs-standard}. The qualitative pattern matches OpenAI: gains concentrate on the entity-sensitive isomorphic dimensions (assignee accuracy, number recall, technical-term recall, speaker consistency) and are systematically compressed by the standard LLM-as-judge scores. The single non-monotonic cell is LLM-judge faithfulness ($-$0.14), which we explain in \S\ref{sec:results:metrics} as a scoring asymmetry against Gemini's workspace-canonical-name expansions rather than a quality regression.

\begin{table}[!t]
  \centering
  \footnotesize
  \setlength{\tabcolsep}{4pt}
  \begin{tabular}{llrrr}
    \toprule
    \textbf{Metric} & \textbf{Type} & \textbf{L1} & \textbf{L3} & \textbf{$\Delta$} \\
    \midrule
    Number recall            & iso. & 0.913 & 0.983 & +0.07 \\
    Technical-term recall    & iso. & 0.756 & 0.802 & +0.05 \\
    AI assignee accuracy     & iso. & 0.101 & 0.666 & +0.57 \\
    Spk.-consistency rate    & iso. & 0.326 & 0.401 & +0.08 \\
    \midrule
    LLM-judge faithfulness   & std. & 3.02  & 2.88  & $-$0.14 \\
    LLM-judge coherence      & std. & 4.74  & 4.88  & +0.14 \\
    AI description quality   & std. & 3.53  & 3.79  & +0.26 \\
    \bottomrule
  \end{tabular}
  \caption{Gemini-stack analog of Table~\ref{tab:isomorphic-vs-standard}: L1$\to$L3 gain on isomorphic latent-state metrics versus standard LLM-as-judge summarization metrics, on a smaller 50-meeting subset of the evaluation data. Same qualitative pattern as OpenAI (isomorphic gains compressed by standard summary-side scores), with a single non-monotonic cell: LLM-judge faithfulness dips slightly at L3 despite uniform improvement on every isomorphic dimension. We explain this as a scoring asymmetry of standard LLM-judge protocols against Gemini's workspace-canonical-name expansions (\S\ref{sec:results:metrics}). Assignee accuracy here is scored under exact match; the OpenAI counterpart in Table~\ref{tab:isomorphic-vs-standard} uses partial-credit substring matching.}
  \label{tab:isomorphic-vs-standard-gemini}
\end{table}

\section{Open-source SpeechBrain baseline}
\label{sec:appendix:speechbrain}

The main-paper evaluation in \S\ref{sec:experiments} focuses on the two state-of-the-art single-vendor systems (OpenAI and Gemini). For reproducibility, we additionally evaluate a fully open-source cascade on a subset of the same evaluation set; we report its system description and per-context-level results here so that the main-paper tables can focus on the state-of-the-art comparison.

\paragraph{System.} The cascade is built around the \texttt{speechbrain} toolkit. CRDNN voice-activity detection (\texttt{vad-crdnn-libriparty}) produces speech regions; wav2vec2 acoustic posteriors (\texttt{asr-wav2vec2-commonvoice-14-en}) are decoded with a KenLM-aware CTC beam search, optionally augmented at L3 with workspace-jargon hot-word biasing through a parallel \texttt{pyctcdecode} \texttt{BeamSearchDecoderCTC} sharing the encoder's KenLM and beam parameters; ECAPA-TDNN embeddings (\texttt{spkrec-ecapa-voxceleb}) feed both spectral-clustering diarization and, at L3, cosine-similarity speaker identification against per-speaker voice enrollments. Prompted summarization, action-item extraction, and outline generation are then performed on the resulting speaker-labeled transcript by a text LLM (\texttt{claude-haiku-4-5}). The cascade uses the same level-conditioned CONTEXT block as the OpenAI and Gemini systems (\S\ref{sec:experiments:context}); L3 additionally enables the hot-word biasing and enrollment-based SID stages described above.

\paragraph{Results.} Table~\ref{tab:speechbrain-headline} reports the SpeechBrain cascade's per-context-level numbers on the same metric slice as the state-of-the-art systems in Table~\ref{tab:headline}. WER is roughly $4\times$ higher than the OpenAI stack across all three context levels, and the cascade trails both state-of-the-art systems on every metric reported here. The L3 increment that monotonically helps OpenAI and Gemini does not transfer cleanly to this cascade: enrollment-driven metrics (cross-meeting speaker-identity consistency and action-item assignee accuracy) jump sharply at L3 once SID is enabled, and number recall and action-item description quality also improve, but WER degrades from $0.557$ to $0.638$ and technical-term recall regresses from $0.361$ to $0.302$. We do not draw any context-sensitivity claims from these numbers; the cascade is included only as an open reference point that future work can re-run end-to-end without commercial API access.

\paragraph{Why L3 hurts WER and term recall in the cascade.} The L3 vocabulary-correction pass is implemented as a constrained LLM rewrite over the upstream ASR transcript, prompted to fix workspace-jargon spellings while explicitly preserving non-jargon content (\emph{``do not paraphrase, reorder, or summarize''}). On the OpenAI and Gemini stacks this prompt sees nearly clean input (WER $\sim$0.10) and the rewrite is essentially a vocabulary patch. On SpeechBrain it sees input at WER $\sim$0.56, which the rewriter treats as noise and silently compresses around: across 10 evaluation meetings (409 turns), the mean L3/L1 word-count ratio per turn is 0.75 (i.e., the rewriter deletes $\sim$25\% of words on average), 31\% of turns lose more than 30\% of their input words, and 13\% lose more than 50\%. Because technical-term recall is computed as a substring scan over the rewritten transcript, every deleted ground-truth term costs a hit; WER moves the same direction because deleted reference-aligned words score as deletions. The same pass on stronger upstream ASR is essentially a no-op, which is why the corresponding L3 cells in Table~\ref{tab:headline} do not show this regression. This points to a deployment-level lesson rather than a benchmark artifact: prompted post-correction passes that are safe at low WER can become silent-summarization stages at high WER, and cascades that pair a weak ASR front-end with an LLM post-corrector should either gate the rewrite on per-segment confidence or replace it with a token-level constrained edit. We leave that to future work and report the unmitigated cascade numbers here.

\paragraph{Qualitative example.} A representative turn from the SpeechBrain L1 transcript and its L3 vocabulary-corrected counterpart on the same audio shows the failure mode concretely:

\begin{quote}\small
\textbf{L1 (raw SpeechBrain):} \emph{``RACE CONDITIONS ISOLATED IT S THE LOCKING NOW REGISTRY CAN CURRENT WRITES GLOBAL LOCKS OUT P NINETY NINE WAS LIKE ONE TWENTY DEVONSONCIS \ldots''}

\textbf{L3 (after vocab-correction LLM):} \emph{``RACE CONDITIONS ISOLATED LOCKING NOW REGISTRY GLOBAL LOCKS OUT P99 LIKE TWENTY DEVON \ldots''}
\end{quote}

The rewriter's intended behavior is visible (\textsc{P NINETY NINE} $\to$ \textsc{P99}, a workspace-jargon recovery), but it co-occurs with the silent compression mode the appendix is documenting: function-word spans (\emph{IT S THE}, \emph{CAN CURRENT WRITES}, \emph{WAS}, \emph{ONE}) and a partially garbled name (\textsc{DEVONSONCIS} $\to$ \textsc{DEVON}) are removed without flagging. Because the upstream WER is already $\sim$0.56, the rewriter cannot tell garbled-but-recoverable spans from genuine acoustic noise and discards both, so every deleted reference-aligned word becomes a deletion error in WER and a missed hit in technical-term recall. The same prompt over OpenAI's $\sim$0.10-WER input has nothing comparable to discard.

\begin{table}[!t]
  \centering
  \small
  \setlength{\tabcolsep}{4pt}
  \begin{tabular}{lrrr}
    \toprule
    \textbf{Metric} & \textbf{L1} & \textbf{L2} & \textbf{L3} \\
    \midrule
    WER~$\downarrow$                  & 0.559 & 0.557 & 0.638 \\
    Number recall~$\uparrow$          & 0.259 & 0.259 & 0.397 \\
    Technical-term recall~$\uparrow$  & 0.364 & 0.361 & 0.302 \\
    AI description quality~$\uparrow$ & 2.08  & 1.88  & 2.13  \\
    Assignee accuracy~$\uparrow$      & 0.012 & 0.017 & 0.465 \\
    Speaker consistency~$\uparrow$    & 0.244 & 0.215 & 0.478 \\
    \bottomrule
  \end{tabular}
  \caption{Open-source SpeechBrain cascade on a 27-meeting single-workspace subset of the evaluation set used in Table~\ref{tab:headline}, and reported as a reproducible open-source reference point only; the subset is smaller than the 238-meeting main evaluation, so these numbers are not directly comparable cell-for-cell. The cascade trails both state-of-the-art systems on every metric reported here. Enrollment-driven metrics (assignee accuracy, speaker consistency) jump sharply at L3 once SID is enabled, but WER and technical-term recall regress at L3 because the L3 vocabulary-correction LLM pass silently summarizes around the cascade's high-WER upstream output (see prose above).}
  \label{tab:speechbrain-headline}
\end{table}

\section{Robustness to the choice of judge}
\label{sec:appendix:qwen-judge}

The judged metrics in \S\ref{sec:experiments:metrics} use \texttt{claude-sonnet-4-6}, which is also the family used to generate the corpus (\S\ref{sec:dataset:pipeline}). To check that the reported context effect is not an artifact of that overlap, we re-ran the judged metrics over the OpenAI system's outputs on the same 238-meeting evaluation set, with the same prompts and an open-weights judge, Qwen3-235B-Instruct \citep{qwen3-2025} served locally through vLLM (Table~\ref{tab:qwen-judge}). Faithfulness and action-item description quality improve from L1 to L3 as they do under the Claude judge. Coherence does not: under this prompt the Qwen judge awards 5.0 in the large majority of cases, leaving no headroom, so the $+0.01$ difference reflects saturation rather than an absence of improvement. We read this as evidence that the L1$\to$L3 improvement does not depend on the judge, whereas the range and calibration of the assigned scores is model-family dependent.

\begin{table}[!t]
  \centering
  \footnotesize
  \setlength{\tabcolsep}{2pt}
  \begin{tabular}{lrrr}
    \toprule
    \textbf{Metric} & \textbf{L1} & \textbf{L3} & \textbf{$\Delta$} \\
    \midrule
    Faithfulness           & 4.00\pmci{.10} & 4.43\pmci{.10} & +0.43 \\
    Coherence              & 4.99\pmci{.02} & 5.00\pmci{.00} & +0.01 \\
    AI description quality & 3.87\pmci{.05} & 4.05\pmci{.04} & +0.18 \\
    \bottomrule
  \end{tabular}
  \caption{Judged metrics recomputed with Qwen3-235B-Instruct as the judge, on the same 238-meeting evaluation set and with the same prompts as Table~\ref{tab:isomorphic-vs-standard} (95\% bootstrap percentile interval, 1{,}000 resamples). The direction of the L1$\to$L3 effect is preserved; coherence is saturated at the 5.00 ceiling under this prompt, which also truncates its interval.}
  \label{tab:qwen-judge}
\end{table}

\section{Licenses of open-source models and datasets}
\label{sec:appendix:licenses}

Table~\ref{tab:licenses} lists the licenses of the open-source models, software packages, and datasets used in the construction and evaluation of \textsc{OSynth}. All assets are used in a manner consistent with their stated licenses and intended use. The released \textsc{OSynth} corpus itself contains no derivatives of the audio content of these datasets: Common Voice clips are used solely as short voice-cloning prompts that condition Qwen3-TTS \citep{qwen3-tts-2025} at synthesis time, and AMI \citep{carletta-etal-2005-ami} and NOTSOFAR-1 \citep{vinnikov-etal-2024-notsofar} excerpts appear only as the real-meeting side of the human-discrimination study (\S\ref{sec:dataset:realism}) and, in the case of AMI, as a style reference to the adversarial rewriter (\S\ref{sec:dataset:refinement}).

For the Common Voice data, we use the English subset of the Mozilla Common Voice \emph{Spontaneous Speech} corpus, version 3.0 (\texttt{sps-corpus-3.0}), chosen over the larger \emph{Scripted Speech} releases because its conversational, unscripted prompts yield voice-cloning references whose prosody better matches the spontaneous register of workspace meetings. The corpus is distributed under CC0 1.0 \citep{ardila-etal-2020-common-voice} and, as required by the Mozilla Data Collective terms, we do not attempt to identify speakers and propagate this restriction to downstream users via the \textsc{OSynth} release notes.

\begin{table*}[!t]
  \centering
  \footnotesize
  \setlength{\tabcolsep}{5pt}
  \begin{tabular}{p{5.0cm}llp{6.5cm}}
    \toprule
    \textbf{Asset} & \textbf{Type} & \textbf{License} & \textbf{Use in \textsc{OSynth}} \\
    \midrule
    Qwen3-TTS \citep{qwen3-tts-2025}                                    & Model    & Apache 2.0 & Multi-speaker TTS with voice cloning (\S\ref{sec:dataset:rendering}) \\
    SpeechBrain toolkit                                                 & Software & Apache 2.0 & Open-source cascade baseline (Appendix~\ref{sec:appendix:speechbrain}) \\
    \texttt{vad-crdnn-libriparty}                                       & Model    & Apache 2.0 & Voice-activity detection in the cascade \\
    \texttt{asr-wav2vec2-commonvoice-14-en}                             & Model    & Apache 2.0 & Acoustic posteriors in the cascade \\
    \texttt{spkrec-ecapa-voxceleb}                                      & Model    & Apache 2.0 & Diarization and SID in the cascade \\
    \texttt{pyctcdecode}                                                & Software & Apache 2.0 & CTC beam-search decoding with hot-word biasing (L3) \\
    \texttt{pyroomacoustics} \citep{scheibler-etal-2018-pyroomacoustics} & Software & MIT        & Room-acoustics post-processing (Appendix~\ref{sec:appendix:audio-postproc}) \\
    \texttt{meeteval}                                                   & Software & MIT        & \textsc{WER} variants in the released evaluation harness \\
    \midrule
    Common Voice Spontaneous Speech 3.0 -- English \citep{ardila-etal-2020-common-voice} & Dataset & CC0 1.0  & Voice-cloning reference clips per persona (\S\ref{sec:dataset:rendering}) \\
    AMI Meeting Corpus \citep{carletta-etal-2005-ami}                   & Dataset  & CC BY 4.0  & Style reference for the adversarial rewriter and real side of the 2AFC and LLM-judge realism studies (\S\ref{sec:dataset:refinement}, \S\ref{sec:dataset:realism}) \\
    NOTSOFAR-1 \citep{vinnikov-etal-2024-notsofar}                      & Dataset  & CC BY 4.0  & Real side of the 2AFC realism study (\S\ref{sec:dataset:realism}) \\
    \bottomrule
  \end{tabular}
  \caption{Licenses of the open-source models, software packages, and datasets used in the construction and evaluation of \textsc{OSynth}. SpeechBrain pretrained checkpoints listed individually inherit the Apache 2.0 license of the SpeechBrain toolkit and are distributed via the corresponding Hugging Face repositories. For NOTSOFAR-1, we follow the CC BY 4.0 license stated in the official GitHub repository.}
  \label{tab:licenses}
\end{table*}

\end{document}